\documentclass[twocolumn]{article}
\IfFileExists{newtxtext.sty}{
  \usepackage{newtxtext,newtxmath}
}{
  \usepackage[T1]{fontenc}
  \usepackage{lmodern}
}
\usepackage[round]{natbib}
\usepackage{geometry}
\usepackage{amsmath,amsfonts}
\usepackage{graphicx}
\usepackage{booktabs}
\usepackage{multirow}
\usepackage{xcolor}
\usepackage{algorithm}
\usepackage{algorithmic}
\usepackage[colorlinks,citecolor=blue,linkcolor=blue,urlcolor=blue]{hyperref}
\usepackage{subcaption}
\usepackage{enumitem}
\usepackage{placeins}
\usepackage{tikz}
\usetikzlibrary{positioning,arrows.meta,calc,patterns,decorations.pathreplacing}

\newcommand{\method}{CAFlow}
\newcommand{\ours}{\method}

\newcommand{\ie}{i.e.}

\newcommand{\Real}{\mathbb{R}}

\begin{document}

\title{CAFlow: Adaptive-Depth Single-Step Flow Matching for Efficient Histopathology Super-Resolution}
\author{Elad Yoshai \and Ariel D. Yoshai \and Natan T. Shaked\thanks{Corresponding author. Email: nshaked@tau.ac.il}\\[4pt]
School of Biomedical Engineering, Tel Aviv University, Tel Aviv, Israel}
\date{}
\maketitle

\begin{abstract}
In digital pathology, whole-slide images routinely exceed gigapixel resolution, making computationally intensive deep generative super-resolution (SR) impractical for routine deployment.
Flow matching models achieve strong SR quality, but they apply the same network depth to every image, wasting computation on inputs that converge at shallow depth.
We introduce \ours{}, an adaptive-depth flow matching framework that routes each image tile to the shallowest network exit that preserves reconstruction quality.
\ours{} performs single-step flow matching in pixel-unshuffled rearranged space, reducing spatial computation by $16\times$ while enabling direct inference.
We show that dedicating half of training to exact $t{=}0$ samples is essential for single-step quality ($-$1.5\,dB without it), a finding with practical implications for deployment since inference always operates at this timestep.
The backbone, FlowResNet (1.90M parameters), mixes convolution and window self-attention blocks across four early exits spanning 3.1 to 13.3\,GFLOPs.
A lightweight exit classifier (${\sim}$6K parameters) routes each tile to the shallowest exit that preserves quality, achieving 33\% compute savings at only 0.12\,dB cost.
We validate clinical relevance through downstream nuclei segmentation with StarDist~\citep{schmidt2018stardist}, confirming that SR quality preserves detection F1 score.
We also show that spatially coherent whole-slide routing correlates with tissue complexity without explicit tissue-type supervision.
On multi-organ histopathology $\times$4 SR, adaptive routing achieves 31.72\,dB PSNR at 33\% compute savings versus full-depth (31.84\,dB), while the shallowest exit alone exceeds bicubic by $+$1.9\,dB at 2.8$\times$ less compute than SwinIR-light~\citep{liang2021swinir}.
The method generalizes to held-out colon tissue with minimal quality loss ($-$0.02\,dB), and at $\times$8 upscaling it outperforms all comparable-compute baselines while remaining competitive with the much larger SwinIR-Medium model.
The model trains in under 5 hours on a single GPU, and adaptive routing can reduce whole-slide inference from minutes to seconds, making generative SR practical for high-throughput pathology workflows.
\end{abstract}

\section{Introduction}
\label{sec:intro}

Single-image super-resolution (SISR) is a fundamental computer vision task with particular importance in medical imaging, where high-resolution (HR) detail can be critical for diagnosis~\citep{wang2020deep}. Computational pathology has become increasingly important for clinical workflows~\citep{abels2019computational,baxi2024digital}.
In computational pathology, tissue specimens captured at lower magnifications or through frozen-section protocols often lack the fine-grained detail of permanent-section images, motivating $\times 4$ super-resolution as a practical tool for improving diagnostic quality~\citep{rivenson2019virtual,de2021deep,yoshai2024super}.

Recent advances in generative modeling have introduced flow matching~\citep{lipman2023flow,liu2023flow} as a powerful framework for image restoration.
Unlike diffusion models~\citep{ho2020denoising,song2021scorebased} that require complex noise schedules, flow matching learns a velocity field that transports samples along straight-line paths from degraded to clean images, enabling efficient single-step inference~\citep{lipman2023flow,albergo2023building}.

However, existing flow-matching SR methods apply uniform network depth to every image.
This is wasteful: some images (smooth tissue backgrounds, homogeneous regions) converge after only a few residual blocks, while others (dense nuclei clusters, fine cellular structures) benefit from the full network depth including expensive attention layers.
Applying the same computation to all images means that easy inputs receive unnecessary processing, inflating inference cost without improving quality.

We address this with \ours{} (Compute-Adaptive Flow matching), a framework with three contributions:

\begin{enumerate}[leftmargin=*,topsep=2pt,itemsep=1pt]
\item \textbf{Single-step flow matching in rearranged space.}
Existing flow matching SR methods operate either in Variational Auto-Encoder (VAE) latent space~\citep{wang2024exploiting} or directly at full high-resolution (HR) resolution~\citep{saharia2022image}, incurring high memory and compute costs.
We apply pixel-unshuffle~\citep{shi2016real} to map both the degraded and target images into a compact representation with $3s^2$ channels at $\frac{1}{s}$ the spatial resolution in each dimension, reducing the spatial compute of all network operations by $s^2{=}16\times$.
Crucially, we show that dedicating 50\% of training to exact $t{=}0$ samples (where $t$ denotes the flow matching timestep that interpolates between the degraded input at $t{=}0$ and the target at $t{=}1$) is essential for single-step inference quality ($-$1.54\,dB without it), a finding with practical implications for deployment since inference always evaluates the model at $t{=}0$.
In practice, this means that throughout training, half of the samples in each mini-batch are drawn at exact $t{=}0$, with the remaining half sampled from the standard continuous timestep distribution.

\item \textbf{Graduated hybrid backbone with quality-aware routing.}
FlowResNet is a 16-block network mixing FiLMResBlocks (cheap convolutions with Feature-wise Linear Modulation (FiLM) conditioning~\citep{perez2018film}) and HybridFiLMBlocks (convolution plus window self-attention~\citep{liu2021swin,liang2021swinir}).
Hybrid blocks are concentrated in later exits, creating a graduated compute schedule from 3.1 to 13.3 giga floating-point operations (GFLOPs) (4.3$\times$ ratio).
A lightweight ExitClassifier (${\sim}$6K parameters) predicts the optimal exit from early backbone features, trained on oracle labels derived from per-exit reconstruction quality.
At inference, each image is routed to its predicted exit via a single \texttt{argmax}, achieving 33\% compute savings at 0.12\,dB Peak Signal-to-Noise Ratio (PSNR) cost.

\item \textbf{Quantitative downstream validation and generalization.}
We validate that SR quality translates to preserved performance in automated downstream tasks: nuclei detection using StarDist~\citep{schmidt2018stardist} on SR outputs preserves detection F1 score compared to HR ground truth, confirming that the reconstruction quality of \ours{} is sufficient for quantitative computational pathology workflows.
Per-organ analysis across breast, kidney, and lung tissue reveals organ-specific performance variation, and whole-slide routing correlates with tissue complexity without explicit tissue-type supervision.
We further demonstrate cross-organ generalization to held-out colon tissue and scaling to $\times$8 upscaling, where \ours{}'s advantage over baselines increases.
\end{enumerate}

On multi-organ histopathology $\times 4$ SR (breast, kidney, lung), adaptive routing achieves 31.72\,dB at 8.9\,GFLOPs (33\% savings vs.\ full-depth 31.84\,dB), while Exit~0 alone exceeds bicubic by $+$1.9\,dB at $2.8\times$ less compute than SwinIR-light~\citep{liang2021swinir}.

\section{Related Work}
\label{sec:related}

\paragraph{Super-Resolution in Digital Pathology}
Super-resolution has emerged as a practical tool in digital pathology, enabling improved visualization of tissue microstructure from lower-magnification or frozen-section acquisitions~\citep{rivenson2019virtual,de2021deep,yoshai2024super}.
While U-Net-based architectures~\citep{ronneberger2015unet} have been foundational in medical image analysis, the computational cost of modern deep generative SR models remains a barrier to large-scale adoption, particularly for whole-slide images containing hundreds to thousands of tiles per specimen.

\paragraph{Single Image Super-Resolution}
Deep learning approaches to SISR have progressed from Convolutional Neural Networks (CNNs)~\citep{he2016deep} (SRCNN~\citep{dong2014learning}, EDSR~\citep{lim2017enhanced}, RCAN~\citep{zhang2018image}) to transformer-based models (SwinIR~\citep{liang2021swinir}, SRFormer~\citep{zhou2023srformer}).
Hybrid conv\,+\,attention architectures such as Restormer~\citep{zamir2022restormer} and NAFNet~\citep{chen2022nafnet} combine local convolutional processing with global attention for image restoration.
Generative approaches include Generative Adversarial Networks (GAN) based methods (ESRGAN~\citep{wang2018esrgan}, Real-ESRGAN~\citep{wang2021real}), diffusion-based methods (SR3~\citep{saharia2022image}, SRDiff~\citep{li2022srdiff}, StableSR~\citep{wang2024exploiting}), and flow-based methods~\citep{liang2021hierarchical}.
All these approaches apply uniform computation across all images.

\paragraph{Flow Matching and Continuous Normalizing Flows}
Flow matching~\citep{lipman2023flow} and rectified flow~\citep{liu2023flow} learn a velocity field to transport distributions along optimal transport paths.
Conditional flow matching~\citep{tong2024improving} enables flexible conditioning mechanisms.
These methods offer advantages over diffusion models in training stability and generation efficiency~\citep{albergo2023building}. A comprehensive treatment of flow matching theory and practice is provided by \citet{lipman2024flow}.
However, existing flow matching methods use fixed-depth networks without adaptive computation.

\paragraph{Adaptive Computation and Early Exit}
Spatially adaptive computation has been explored through stochastic depth~\citep{huang2016deep}, early-exit networks~\citep{teerapittayanon2016branchynet}, mixture of experts~\citep{shazeer2017outrageously}, and spatially adaptive normalization (SPADE~\citep{park2019semantic}).
DPM-Solver~\citep{lu2022dpm} uses adaptive step sizes globally but not per image.
To our knowledge, \ours{} is the first work to combine early-exit routing with flow matching for super-resolution.
Unlike prior early-exit approaches that rely on confidence thresholds or fixed exit schedules, our quality-aware classifier is trained on oracle labels derived from per-exit reconstruction quality, enabling principled per-image routing without ground truth at inference.
Critically, diffusion-based SR methods such as SR3~\citep{saharia2022image} require iterative multi-step denoising where each step conditions on the output of the previous, making per-image depth adaptation fundamentally difficult: all steps must complete before any output is available.
Flow matching's single-step formulation uniquely enables early-exit routing because the entire velocity prediction is produced in one forward pass through a graduated backbone, allowing the model to exit at any depth without sacrificing the iterative structure.

\paragraph{Window Self-Attention}
Building on the multi-head self-attention mechanism~\citep{vaswani2017attention} and its adaptation to vision~\citep{dosovitskiy2021image}, the Swin Transformer~\citep{liu2021swin} introduced Window Multi-head Self-Attention (W-MSA) and its Shifted-Window variant (SW-MSA) for efficient vision processing.
SwinIR~\citep{liang2021swinir} adapted this for image restoration.
Our HybridFiLMBlock integrates window attention into FiLM-conditioned residual blocks, concentrating attention in later exits where global receptive field most benefits quality.

\paragraph{Loss-Aware and Adaptive Timestep Sampling}
For diffusion models, P2 weighting~\citep{choi2022perception} and Min-SNR~\citep{hang2024efficient} adapt loss weights across timesteps, but these are global schedules.
Curriculum learning~\citep{bengio2009curriculum} and self-paced learning~\citep{kumar2010self} schedule training examples by difficulty.
Our loss-aware timestep sampler builds on these ideas with a simple exponential moving average (EMA)-based approach that concentrates gradient signal on difficult timesteps without additional neural networks.

\section{Method}
\label{sec:method}

\subsection{Preliminaries: Flow Matching in Rearranged Space}
\label{sec:prelim}

Given a low-resolution (LR) image $\mathbf{x}_{LR} \in \Real^{3 \times H \times W}$, we define the starting point $\mathbf{x}_0 = \text{Bicubic}_{\uparrow s}(\mathbf{x}_{LR})$ and the target $\mathbf{x}_1 = \mathbf{x}_{HR}$, where $s{=}4$ is the scale factor.

Rather than operating in pixel space, we apply pixel-unshuffle~\citep{shi2016real} to map images into a compact rearranged representation:
\begin{equation}
\tilde{\mathbf{x}} = \text{PixelUnshuffle}_s(\mathbf{x}) \in \Real^{3s^2 \times \frac{H}{s} \times \frac{W}{s}}.
\label{eq:rearrange}
\end{equation}
For $s{=}4$ and $256{\times}256$ crops, this yields 48-channel feature maps at $64{\times}64$ spatial resolution.
This rearrangement reduces spatial dimensions by $s^2$, enabling larger batch sizes and avoiding sub-pixel upsampling layers, while being exactly invertible via pixel-shuffle.
All flow matching operates in this rearranged space; we omit the tilde in subsequent equations for brevity, reintroducing it in the algorithms where the rearrangement is made explicit.

The linear interpolation path and target velocity are:
\begin{equation}
\mathbf{x}_t = (1{-}t)\,\mathbf{x}_0 + t\,\mathbf{x}_1, \quad \mathbf{v} = \mathbf{x}_1 - \mathbf{x}_0, \quad t \in [0, 1].
\label{eq:interpolation}
\end{equation}
A neural network $\mathbf{v}_\theta$ is trained to predict the velocity:
\begin{equation}
\mathcal{L}_{\text{vel}} = \mathbb{E}_{t \sim p(t)} \big[\, \|\mathbf{v}_\theta(\mathbf{x}_t, \mathbf{x}_0, t) - \mathbf{v}\|_1 \,\big],
\label{eq:fm_loss}
\end{equation}
where the timestep distribution $p(t)$ is described in Section~\ref{sec:timestep}.
At inference, we evaluate only at $t{=}0$, directly predicting the residual from bicubic-upsampled LR to HR in a single forward pass.
At inference, a single forward pass at $t{=}0$ yields:
\begin{equation}
\hat{\mathbf{x}}_1 = \mathbf{x}_0 + \mathbf{v}_\theta(\mathbf{x}_0, \mathbf{x}_0, 0),
\label{eq:single_step}
\end{equation}
and the HR image is recovered via pixel-shuffle: $\hat{\mathbf{x}}_{HR} = \text{PixelShuffle}_s(\hat{\mathbf{x}}_1)$.
In principle, multi-step Euler integration ($\mathbf{x}_{t+\Delta t} = \mathbf{x}_t + \Delta t \cdot \mathbf{v}_\theta$) can also be applied when the velocity field is sufficiently smooth.

\subsection{Hybrid Backbone: FlowResNet}
\label{sec:backbone}

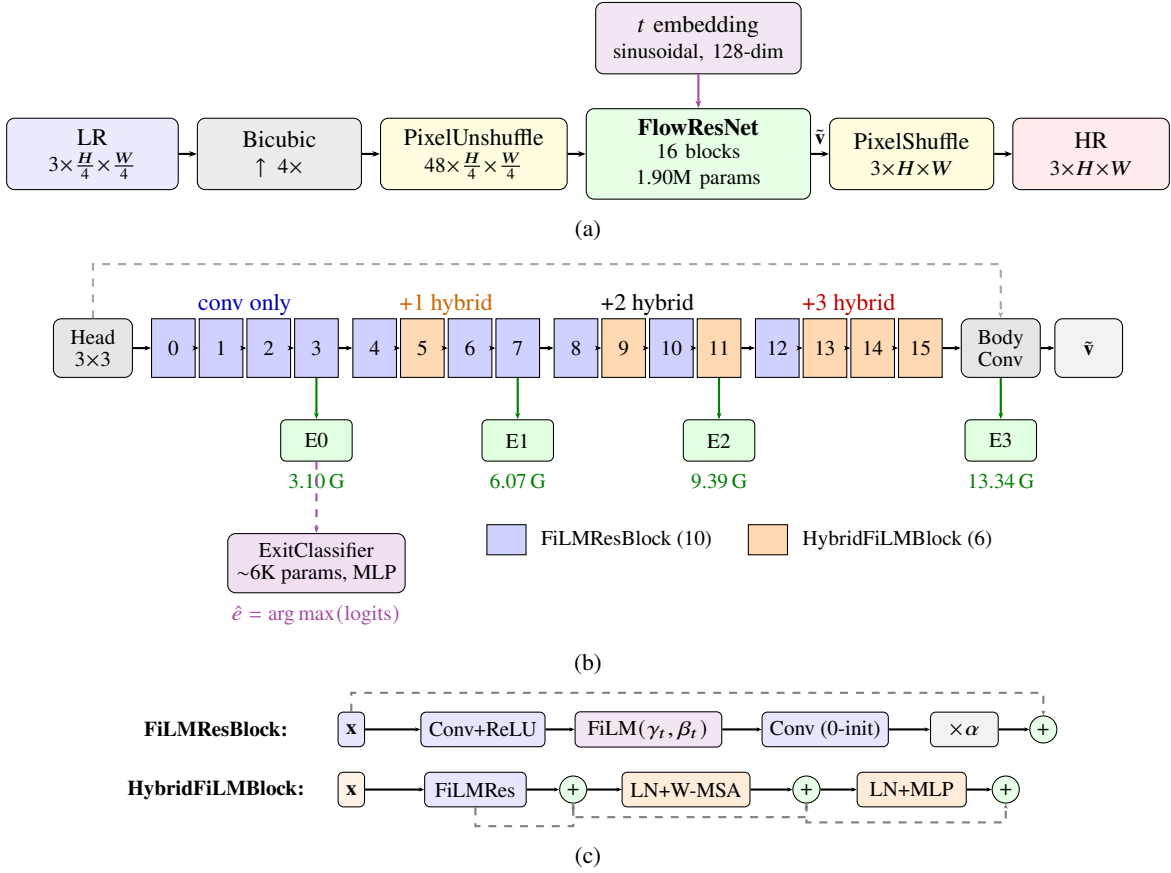
\begin{figure*}[!t]
\centering

\begin{subfigure}[b]{\textwidth}
\centering
\begin{tikzpicture}[
  box/.style={draw, rounded corners=3pt, minimum height=0.95cm, inner sep=4pt, font=\small, align=center, text width=2.0cm},
  arr/.style={-{Stealth[length=3pt]}, thick},
]
\node[box, fill=blue!8] (lr) {LR\\[-1pt]{\footnotesize $3{\times}\frac{H}{4}{\times}\frac{W}{4}$}};
\node[box, fill=gray!15, text width=1.9cm, right=0.24cm of lr] (bic) {Bicubic\\[-1pt]{\footnotesize $\uparrow 4{\times}$}};
\node[box, fill=yellow!15, text width=2.2cm, right=0.24cm of bic] (unshuffle) {PixelUnshuffle\\[-1pt]{\footnotesize $48{\times}\frac{H}{4}{\times}\frac{W}{4}$}};
\node[box, fill=green!10, text width=2.7cm, right=0.24cm of unshuffle] (backbone) {\textbf{FlowResNet}\\[-1pt]{\footnotesize 16 blocks\\[-1pt]1.90M params}};
\node[box, fill=yellow!15, text width=1.9cm, right=0.24cm of backbone] (shuffle) {PixelShuffle\\[-1pt]{\footnotesize $3{\times}H{\times}W$}};
\node[box, fill=red!8, text width=1.8cm, right=0.24cm of shuffle] (hr) {HR\\[-1pt]{\footnotesize $3{\times}H{\times}W$}};

\node[box, fill=violet!10, text width=2.45cm, above=0.45cm of backbone] (temb) {$t$ embedding\\[-1pt]{\footnotesize sinusoidal, 128-dim}};
\draw[arr, violet!70] (temb) -- (backbone);

\draw[arr] (lr) -- (bic);
\draw[arr] (bic) -- (unshuffle);
\draw[arr] (unshuffle) -- (backbone);
\draw[arr] (backbone) -- node[above, font=\footnotesize] {$\tilde{\mathbf{v}}$} (shuffle);
\draw[arr] (shuffle) -- (hr);

\end{tikzpicture}
\caption{}
\label{fig:arch_pipeline}
\end{subfigure}

\vspace{6pt}

\begin{subfigure}[b]{\textwidth}
\centering
\begin{tikzpicture}[
  block/.style={draw, minimum width=0.58cm, minimum height=0.78cm, inner sep=0pt, font=\footnotesize},
  filmblock/.style={block, fill=blue!18},
  hybridblock/.style={block, fill=orange!30},
  exitbox/.style={draw, rounded corners=2pt, fill=green!12, minimum width=0.95cm, minimum height=0.55cm, inner sep=2pt, font=\footnotesize},
  arr/.style={-{Stealth[length=2.5pt]}, thick},
  seg/.style={draw, rounded corners=3pt, dashed, gray!60, inner sep=3pt},
]

\node[draw, rounded corners=3pt, fill=gray!20, minimum width=1.05cm, minimum height=0.78cm, font=\footnotesize, align=center] (head) {Head\\[-1pt]$3{\times}3$};

\node[filmblock, right=0.24cm of head] (b0) {0};
\node[filmblock, right=0.04cm of b0] (b1) {1};
\node[filmblock, right=0.04cm of b1] (b2) {2};
\node[filmblock, right=0.04cm of b2] (b3) {3};

\node[filmblock, right=0.18cm of b3] (b4) {4};
\node[hybridblock, right=0.04cm of b4] (b5) {5};
\node[filmblock, right=0.04cm of b5] (b6) {6};
\node[filmblock, right=0.04cm of b6] (b7) {7};

\node[filmblock, right=0.18cm of b7] (b8) {8};
\node[hybridblock, right=0.04cm of b8] (b9) {9};
\node[filmblock, right=0.04cm of b9] (b10) {10};
\node[hybridblock, right=0.04cm of b10] (b11) {11};

\node[filmblock, right=0.18cm of b11] (b12) {12};
\node[hybridblock, right=0.04cm of b12] (b13) {13};
\node[hybridblock, right=0.04cm of b13] (b14) {14};
\node[hybridblock, right=0.04cm of b14] (b15) {15};

\node[draw, rounded corners=3pt, fill=gray!20, minimum width=1.05cm, minimum height=0.78cm, right=0.24cm of b15, font=\footnotesize, align=center] (body) {Body\\[-1pt]Conv};

\node[draw, rounded corners=3pt, fill=gray!10, minimum width=0.9cm, minimum height=0.78cm, right=0.18cm of body, font=\footnotesize] (out) {$\tilde{\mathbf{v}}$};

\draw[arr] (head) -- (b0);
\foreach \i/\j in {0/1,1/2,2/3}{
  \draw[arr] (b\i) -- (b\j);
}
\draw[arr] (b3) -- (b4);
\foreach \i/\j in {4/5,5/6,6/7}{
  \draw[arr] (b\i) -- (b\j);
}
\draw[arr] (b7) -- (b8);
\foreach \i/\j in {8/9,9/10,10/11}{
  \draw[arr] (b\i) -- (b\j);
}
\draw[arr] (b11) -- (b12);
\foreach \i/\j in {12/13,13/14,14/15}{
  \draw[arr] (b\i) -- (b\j);
}
\draw[arr] (b15) -- (body);
\draw[arr] (body) -- (out);

\draw[arr, dashed, gray!70] (head.north) -- ++(0, 0.72) -| (body.north);

\node[exitbox, below=0.55cm of b3] (e0) {E0};
\draw[arr, green!50!black] (b3.south) -- (e0.north);
\node[font=\footnotesize, below=1pt of e0, green!50!black] {3.10\,G};

\node[exitbox, below=0.55cm of b7] (e1) {E1};
\draw[arr, green!50!black] (b7.south) -- (e1.north);
\node[font=\footnotesize, below=1pt of e1, green!50!black] {6.07\,G};

\node[exitbox, below=0.55cm of b11] (e2) {E2};
\draw[arr, green!50!black] (b11.south) -- (e2.north);
\node[font=\footnotesize, below=1pt of e2, green!50!black] {9.39\,G};

\node[exitbox, below=0.55cm of body] (e3) {E3};
\draw[arr, green!50!black] (body.south) -- (e3.north);
\node[font=\footnotesize, below=1pt of e3, green!50!black] {13.34\,G};

\node[draw, rounded corners=3pt, fill=violet!12, minimum width=2.25cm, minimum height=0.72cm, below=0.95cm of e0, font=\footnotesize, align=center] (cls) {ExitClassifier\\[-1pt]{\footnotesize $\sim$6K params, MLP}};
\draw[arr, violet!70, dashed] (e0.south) -- (cls.north);
\node[font=\footnotesize, below=1pt of cls, violet!70] {$\hat{e} = \arg\max(\mathrm{logits})$};

\node[font=\small, text=blue!70!black, fill=white, inner sep=1pt] at ($(b1)!0.5!(b2)+(0,0.58)$) {conv only};
\node[font=\small, text=orange!80!black, fill=white, inner sep=1pt] at ($(b5)!0.5!(b6)+(0,0.58)$) {+1 hybrid};
\node[font=\small, text=black, fill=white, inner sep=1pt] at ($(b9)!0.5!(b10)+(0,0.58)$) {+2 hybrid};
\node[font=\small, text=red!75!black, fill=white, inner sep=1pt] at ($(b13)!0.5!(b14)+(0,0.58)$) {+3 hybrid};

\node[filmblock, minimum width=0.52cm, minimum height=0.4cm] at ($(b7)+(-0.15, -2.55)$) (leg1) {};
\node[font=\footnotesize, right=2pt of leg1] (leg1txt) {FiLMResBlock (10)};
\node[hybridblock, minimum width=0.52cm, minimum height=0.4cm, right=10pt of leg1txt.east, anchor=west] (leg2) {};
\node[font=\footnotesize, right=2pt of leg2] {HybridFiLMBlock (6)};

\end{tikzpicture}
\caption{}
\label{fig:arch_backbone}
\end{subfigure}

\vspace{4pt}

\begin{subfigure}[b]{\textwidth}
\centering
\begin{tikzpicture}[
  op/.style={draw, rounded corners=2pt, minimum height=0.46cm, inner sep=3pt, font=\footnotesize, align=center},
  arr/.style={-{Stealth[length=2pt]}, thick},
  plus/.style={draw, circle, inner sep=0.5pt, font=\footnotesize, minimum size=0.36cm},
]

\node[font=\footnotesize\bfseries] at (-0.3, 0.6) {FiLMResBlock:};
\node[op, fill=blue!10] (f1) at (1.5, 0.6) {$\mathbf{x}$};
\node[op, fill=blue!10, minimum width=1.65cm] (f2) at (3.25, 0.6) {Conv+ReLU};
\node[op, fill=violet!10, minimum width=1.95cm] (f3) at (5.45, 0.6) {FiLM$(\gamma_t,\beta_t)$};
\node[op, fill=blue!10, minimum width=1.65cm] (f4) at (7.8, 0.6) {Conv (0-init)};
\node[op, fill=gray!10, minimum width=0.9cm] (f5) at (9.65, 0.6) {${\times}\alpha$};
\node[plus, fill=green!10] (fa) at (10.7, 0.6) {$+$};
\draw[arr] (f1) -- (f2); \draw[arr] (f2) -- (f3); \draw[arr] (f3) -- (f4);
\draw[arr] (f4) -- (f5); \draw[arr] (f5) -- (fa);
\draw[arr, dashed, gray] (f1.north) -- ++(0,0.25) -| (fa.north);

\node[font=\footnotesize\bfseries] at (-0.3, -0.2) {HybridFiLMBlock:};
\node[op, fill=orange!10] (h1) at (1.5, -0.2) {$\mathbf{x}$};
\node[op, fill=blue!10, minimum width=1.35cm] (h2) at (3.15, -0.2) {FiLMRes};
\node[plus, fill=green!10] (ha1) at (4.45, -0.2) {$+$};
\node[op, fill=orange!15, minimum width=1.7cm] (h3) at (5.95, -0.2) {LN+W-MSA};
\node[plus, fill=green!10] (ha2) at (7.55, -0.2) {$+$};
\node[op, fill=orange!15, minimum width=1.45cm] (h4) at (8.95, -0.2) {LN+MLP};
\node[plus, fill=green!10] (ha3) at (10.2, -0.2) {$+$};
\draw[arr] (h1) -- (h2); \draw[arr] (h2) -- (ha1); \draw[arr] (ha1) -- (h3);
\draw[arr] (h3) -- (ha2); \draw[arr] (ha2) -- (h4); \draw[arr] (h4) -- (ha3);
\draw[arr, dashed, gray] (h2.south) -- ++(0,-0.25) -| (ha1.south);
\draw[arr, dashed, gray] (ha1.south) -- ++(0,-0.18) -| (ha2.south);
\draw[arr, dashed, gray] (ha2.south) -- ++(0,-0.25) -| (ha3.south);
\end{tikzpicture}
\caption{}
\label{fig:arch_blocks}
\end{subfigure}

\caption{\ours{} architecture overview. (a)~The pipeline operates entirely in pixel-unshuffled rearranged space at $\frac{H}{4}{\times}\frac{W}{4}$ resolution, reducing spatial compute by $16{\times}$. (b)~FlowResNet backbone (1.90M parameters): 16 blocks grouped into 4 exit segments with increasing hybrid (attention) block density, creating a non-linear cost gradient ($4.3{\times}$ ratio). The ExitClassifier (${\sim}$6K parameters) predicts the optimal exit from E0 features. (c)~Internal structure of FiLMResBlock (Eq.~\ref{eq:filmresblock}) and HybridFiLMBlock (Eq.~\ref{eq:hybridblock}). Dashed arrows denote residual connections; HybridFiLMBlock augments the convolutional path with W-MSA/SW-MSA ($w{=}8$, 8 heads) and a feed-forward MLP.}
\label{fig:architecture}
\end{figure*}

The velocity network is FlowResNet, a 16-block residual network~\citep{he2016deep} with two types of blocks arranged in a graduated schedule (Figure~\ref{fig:architecture}).

The base building block follows the Enhanced Deep Residual (EDSR) design~\citep{lim2017enhanced}, a pre-activation residual block without batch normalization that avoids removing range flexibility and reduces GPU memory, augmented with FiLM conditioning~\citep{perez2018film} on the timestep $t$.
A sinusoidal embedding of $t$ is projected to per-channel scale and shift parameters:
\begin{align}
\mathbf{h} &= \text{ReLU}(\text{Conv}_{3\times3}(\mathbf{x})), \nonumber \\
(\boldsymbol{\gamma}, \boldsymbol{\beta}) &= \text{Linear}(\text{SiLU}(\mathbf{e}_t)), \nonumber \\
\mathbf{h} &= \mathbf{h} \odot (1 + \boldsymbol{\gamma}) + \boldsymbol{\beta}, \nonumber \\
\text{FiLMResBlock}(\mathbf{x}) &= \mathbf{x} + \alpha \cdot \text{Conv}_{3\times3}(\mathbf{h}),
\label{eq:filmresblock}
\end{align}
where $\alpha{=}0.1$ is the residual scale and $\mathbf{e}_t$ is the timestep embedding.
The second convolution is zero-initialized so that each block starts as an identity function.

For later exits requiring global context, we augment the FiLMResBlock with window self-attention (W-MSA/SW-MSA~\citep{liu2021swin,liang2021swinir}) to form the HybridFiLMBlock:
W-MSA computes attention within local windows to keep computational cost linear, while SW-MSA shifts these windows in successive layers to build global context.
\begin{align}
\mathbf{x} &\leftarrow \text{FiLMResBlock}(\mathbf{x}, \mathbf{e}_t), \nonumber \\
\mathbf{x} &\leftarrow \mathbf{x} + \text{W-MSA}(\text{LN}(\mathbf{x})), \nonumber \\
\mathbf{x} &\leftarrow \mathbf{x} + \text{MLP}(\text{LN}(\mathbf{x})),
\label{eq:hybridblock}
\end{align}
where LN is LayerNorm~\citep{ba2016layer}, W-MSA uses window size $w{=}8$, 8 attention heads, and learned relative position bias, and the multi-layer perceptron (MLP) has expansion ratio~2.
Consecutive hybrid blocks alternate between standard (shift$=0$) and shifted (shift$=w/2$) windows for cross-window information flow.

As shown in Figure~\ref{fig:architecture}, of the 16 blocks (indices 0 to 15), blocks at indices $\{5, 9, 11, 13, 14, 15\}$ are HybridFiLMBlocks; the remaining 10 are FiLMResBlocks.
Four exit points are placed at evenly-spaced intervals after blocks 3, 7, 11, and 15:
\begin{itemize}[leftmargin=*,topsep=1pt,itemsep=1pt]
\item \textbf{Exit~0 (blocks 0 to 3):} 4 FiLMResBlocks, 0 hybrid blocks, 3.1\,GFLOPs.
\item \textbf{Exit~1 (blocks 0 to 7):} 7 FiLMRes + 1 hybrid, 6.1\,GFLOPs.
\item \textbf{Exit~2 (blocks 0 to 11):} 9 FiLMRes + 3 hybrid, 9.4\,GFLOPs.
\item \textbf{Exit~3 (blocks 0 to 15):} 10 FiLMRes + 6 hybrid, 13.3\,GFLOPs.
\end{itemize}
The concentration of hybrid blocks in later exits creates a non-linear cost gradient: the E3/E0 ratio is 4.3$\times$ rather than the 4.0$\times$ expected from uniform blocks.
Early exits provide local-only processing (cheap, fast), while later exits progressively add global context (expensive, higher quality).

Each intermediate exit (0 to 2) uses a zero-initialized $3{\times}3$ convolution mapping features to $3s^2$-channel velocity in rearranged space.
The final exit (3) uses the global residual connection (\ie, body convolution + tail), giving it access to the full skip connection from the head.

The network input is the 96-channel concatenation $[\tilde{\mathbf{x}}_t;\, \tilde{\mathbf{x}}_0]$ in rearranged space.
The output is a $3s^2{=}48$-channel predicted velocity.
The model totals 1.90M parameters.

\subsection{Quality-Aware Exit Routing}
\label{sec:routing}

At inference, each image should exit at the shallowest depth that preserves quality.
We train a lightweight ExitClassifier to predict the optimal exit from backbone features.

The ExitClassifier is a small MLP applied to features from Exit~0: we perform global average pooling, followed by a fully connected layer ($64 \to 64$) with Rectified Linear Unit (ReLU) activation, a second fully connected layer ($64 \to 32$) with Rectified Linear Unit (ReLU) activation, and a final linear projection ($32 \to 4$) producing logits over the four exits, totaling ${\sim}$6K parameters.
The key design choice is to route from backbone features (after 4 blocks of processing) rather than raw LR input: the network itself reveals whether the image needs further processing.

During training, for samples with $t < 0.15$ (near-inference conditions), we compute the target-image reconstruction L1 loss at each exit:
\begin{equation}
\ell_e = \|\hat{\mathbf{x}}_1^{(e)} - \mathbf{x}_1\|_1, \quad e \in \{0, 1, 2, 3\}.
\label{eq:per_exit_loss}
\end{equation}
The oracle label is the earliest exit whose loss is within $\varepsilon$ of the best:
\begin{equation}
e^* = \min\{e : \ell_e \leq \min_{e'} \ell_{e'} + \varepsilon\}, \quad \varepsilon = 0.02.
\label{eq:oracle}
\end{equation}
This encourages early exit whenever quality is comparable, rather than always selecting the lowest-loss exit (which would bias toward deeper exits).

The classifier is trained with cross-entropy loss on oracle labels. Let $\operatorname{sg}(\cdot)$ denote the stop-gradient operator. Then:
\begin{equation}
\mathcal{L}_{\text{router}} = \text{CE}\!\big(\text{ExitClassifier}(\operatorname{sg}(\mathbf{f}_{E0})),\; e^*\big),
\label{eq:router_loss}
\end{equation}
where ``CE'' denotes the cross-entropy loss function, ``ExitClassifier'' is the lightweight routing network, and $\mathbf{f}_{E0}$ represents the intermediate feature maps extracted at Exit~0. The stop-gradient ensures the router is trained using a separate optimizer and its gradients do not flow into the backbone. This prevents the routing objective from degrading the main network's feature quality. At inference time, the classifier predicts exit logits from E0 features, and each image is routed to $\hat{e} = \arg\max(\text{logits})$. Images predicted for Exit~0 skip all subsequent blocks; images predicted for deeper exits continue processing. This dynamic, quality-aware routing mechanism during deployment is outlined in Algorithm~\ref{alg:inference}.

\subsection{Loss-Aware Timestep Sampling}
\label{sec:timestep}

Standard flow matching samples $t \sim \mathcal{U}(0,1)$, giving equal weight to all timesteps.
We replace this with a strategy that concentrates gradient signal where the model most needs improvement.

We use a logit-normal base distribution:
\begin{equation}
t = \sigma(u), \quad u \sim \mathcal{N}(\mu, \sigma^2_t),
\label{eq:logit_normal}
\end{equation}
with $\mu = -1.0$ and $\sigma_t = 1.0$, concentrating mass on low-to-mid timesteps where inference-relevant predictions occur~\citep{esser2024scaling}.

Since single-step inference evaluates the model at $t{=}0$, we dedicate half of each training batch to exact $t{=}0$ samples, making them equivalent to direct residual prediction.
The remaining 50\% are drawn from the logit-normal or loss-aware distributions.

After warmup, we partition $[0,1]$ into $B = 20$ equal bins and track the EMA of per-bin velocity losses.
After warmup, timesteps are sampled from:
\begin{equation}
p(\text{bin}_b) \propto (1 - \varepsilon)\,\frac{\ell_b^\alpha}{Z} + \frac{\varepsilon}{B},
\label{eq:loss_aware}
\end{equation}
where $\ell_b$ is the EMA loss for bin $b$, $\alpha = 0.3$ controls sharpness, $\varepsilon = 0.5$ is the uniform mixing ratio, and $Z = \sum_b \ell_b^\alpha$.

\subsection{Training Objective and Strategy}
\label{sec:training}

During training, all four exits predict velocities and receive equal weight:
\begin{equation}
\mathcal{L}_{\text{multi}} = \frac{1}{4} \sum_{e=0}^{3} \big( \mathcal{L}_{\text{vel}}^{(e)} + \mathcal{L}_{x_0}^{(e)} \big),
\label{eq:multi_exit_loss}
\end{equation}
where $\mathcal{L}_{\text{vel}}^{(e)} = \|\mathbf{v}_\theta^{(e)} - \mathbf{v}\|_1$ is the velocity L1 loss and $\mathcal{L}_{x_0}^{(e)} = \|\hat{\mathbf{x}}_1^{(e)} - \mathbf{x}_1\|_1$ is the target-image reconstruction loss at exit $e$, with $\hat{\mathbf{x}}_1^{(e)} = \mathbf{x}_t + (1{-}t)\,\mathbf{v}_\theta^{(e)}$.

The Structural Similarity Index Measure (SSIM) loss encourages structural fidelity at the final exit:
\begin{equation}
\mathcal{L}_{\text{SSIM}} = 1 - \text{SSIM}(\hat{\mathbf{x}}_1^{(e_\text{last})}, \mathbf{x}_1).
\end{equation}
The consistency loss enforces that the model predicts the same HR target from different timesteps $t_1, t_2$:
\begin{equation}
\mathcal{L}_{\text{consist}} = \|\hat{\mathbf{x}}_1(t_2) - \operatorname{sg}(\hat{\mathbf{x}}_1(t_1))\|_1.
\end{equation}

The total loss combines these terms:
\begin{equation}
\mathcal{L} = \mathcal{L}_{\text{multi}} + 0.1 \cdot \mathcal{L}_{\text{SSIM}} + 0.1 \cdot \mathcal{L}_{\text{consist}}.
\label{eq:total_loss}
\end{equation}
The router loss $\mathcal{L}_{\text{router}}$ (Eq.~\ref{eq:router_loss}) is optimized with a separate optimizer and does not affect backbone gradients.

Training proceeds in two phases. In Phase~1 (epochs 1 to 5, warmup), standard flow matching is applied with logit-normal $t$ sampling; the router is not yet trained, and per-bin loss statistics are collected.
In Phase~2 (epochs 6 to 700), the router is activated with oracle labels, loss-aware sampling adapts bin probabilities, and all losses are active.
We use batch size~32, $256{\times}256$ random crops, learning rate $2 \times 10^{-4}$ with cosine decay, 8-bit AdamW~\citep{loshchilov2019decoupled}, bfloat16 mixed precision, and EMA decay 0.9999.
The complete end-to-end training procedure is summarized in Algorithm~\ref{alg:training}.

\begin{algorithm}[t]
\caption{\method{} Training}
\label{alg:training}
\begin{algorithmic}[1]
\REQUIRE FlowResNet $\mathbf{v}_\theta$ (4 exits), ExitClassifier $\mathcal{C}$, LossAwareSampler $\mathcal{S}$
\FOR{epoch $= 1, \ldots, 700$}
    \STATE $\texttt{is\_warmup} \leftarrow (\text{epoch} \leq 5)$
    \FOR{each batch $(\mathbf{x}_{LR}, \mathbf{x}_{HR})$}
        \STATE $\tilde{\mathbf{x}}_0 \leftarrow \text{PixelUnshuffle}(\text{Bicubic}_{\uparrow 4}(\mathbf{x}_{LR}))$
        \STATE $\tilde{\mathbf{x}}_1 \leftarrow \text{PixelUnshuffle}(\mathbf{x}_{HR})$
        \STATE $t \leftarrow \mathcal{S}.\text{sample}(\text{epoch})$ \COMMENT{logit-normal or loss-aware}
        \STATE $\tilde{\mathbf{x}}_t \leftarrow (1{-}t)\,\tilde{\mathbf{x}}_0 + t\,\tilde{\mathbf{x}}_1$
        \STATE $[\mathbf{v}^{(0)}, \ldots, \mathbf{v}^{(3)}], \text{logits}, \text{feats} \leftarrow \mathbf{v}_\theta(\tilde{\mathbf{x}}_t, \tilde{\mathbf{x}}_0, t)$
        \STATE $\mathcal{L} \leftarrow \frac{1}{4}\sum_{e}(\mathcal{L}_{\text{vel}}^{(e)} + \mathcal{L}_{x_0}^{(e)}) + 0.1\,\mathcal{L}_{\text{SSIM}} + 0.1\,\mathcal{L}_{\text{consist}}$
        \STATE Update $\theta$ with $\mathcal{L}$; update $\mathcal{S}$ with per-bin losses
        \IF{not is\_warmup \AND $t < 0.15$}
            \STATE $e^* \leftarrow \min\{e : \ell_e \leq \min_{e'}\ell_{e'} + \varepsilon\}$ \COMMENT{Oracle label}
            \STATE $\mathcal{L}_{\text{router}} \leftarrow \text{CE}(\text{logits}, e^*)$
            \STATE Update $\mathcal{C}$ with $\mathcal{L}_{\text{router}}$ \COMMENT{Separate optimizer}
        \ENDIF
    \ENDFOR
\ENDFOR
\end{algorithmic}
\end{algorithm}

\begin{algorithm}[t]
\caption{\method{} Adaptive Inference}
\label{alg:inference}
\begin{algorithmic}[1]
\REQUIRE LR image $\mathbf{x}_{LR}$, FlowResNet $\mathbf{v}_\theta$, ExitClassifier $\mathcal{C}$
\STATE $\tilde{\mathbf{x}}_0 \leftarrow \text{PixelUnshuffle}(\text{Bicubic}_{\uparrow 4}(\mathbf{x}_{LR}))$
\STATE Process blocks 0 to 3 to obtain E0 features $\mathbf{f}_{E0}$
\STATE $\hat{e} \leftarrow \arg\max\!\big(\mathcal{C}(\mathbf{f}_{E0})\big)$ \COMMENT{Predict optimal exit}
\IF{$\hat{e} = 0$}
    \STATE $\tilde{\mathbf{v}} \leftarrow \text{ExitHead}_0(\mathbf{f}_{E0})$ \COMMENT{Skip blocks 4 to 15}
\ELSE
    \STATE Continue processing blocks $4, \ldots$ up to exit $\hat{e}$
    \STATE $\tilde{\mathbf{v}} \leftarrow$ velocity at exit $\hat{e}$
\ENDIF
\STATE $\hat{\mathbf{x}}_{HR} \leftarrow \text{PixelShuffle}\!\big(\tilde{\mathbf{x}}_0 + \tilde{\mathbf{v}}\big)$
\RETURN $\hat{\mathbf{x}}_{HR}$
\end{algorithmic}
\end{algorithm}

\section{Experiments}
\label{sec:experiments}

\subsection{Experimental Setup}

We use multi-organ histopathology patches from The Cancer Genome Atlas (TCGA), comprising breast, kidney, and lung tissue.
The dataset contains 3,090 training and 343 validation patches at $1024 \times 1024$ pixels, including both frozen and permanent section types.
During training, we extract random $256 \times 256$ crops with horizontal/vertical flip, rotation, and mild color jitter (brightness=0.1, contrast=0.1, saturation=0.05).
LR images are generated via $\times 4$ bicubic downsampling.

We compare against the following baselines:
Bicubic interpolation;
EDSR~\citep{lim2017enhanced} (1.52M parameters, 16.25\,GFLOPs) with sub-pixel upsampling;
SwinIR-light~\citep{liang2021swinir} (0.93M parameters, 8.72\,GFLOPs) using the BasicSR implementation;
SwinIR-Medium~\citep{liang2021swinir} (11.90M parameters, 107.11\,GFLOPs), trained from scratch with the same protocol;
SRFormer-light~\citep{zhou2023srformer} (0.87M parameters, 8.25\,GFLOPs), a recent transformer baseline using permuted self-attention with window size~16;
and SR3~\citep{saharia2022image}, a Denoising Diffusion Probabilistic Models (DDPM)-based diffusion SR model (2.94M parameters) with SDEdit-style initialization, 10 Denoising Diffusion Implicit Models (DDIM) steps $\times$ 4 averaged samples, 4{,}376 total GFLOPs.
Regression baselines (EDSR, SwinIR-light, SwinIR-Medium, SRFormer-light) process at LR resolution with PixelShuffle~\citep{shi2016real} upsampling; SR3 operates at HR resolution (256$\times$256).
All baselines are trained from scratch on our dataset with the same optimizer settings.

We report Peak Signal-to-Noise Ratio (PSNR, $\uparrow$), SSIM~\citep{wang2004image} ($\uparrow$), and Learned Perceptual Image Patch Similarity (LPIPS, $\downarrow$)~\citep{zhang2018unreasonable}.
GFLOPs are measured using PyTorch utilities (\texttt{FlopCounterMode}) at $64{\times}64$ LR resolution.

All models are implemented in PyTorch 2.10, RTX~5070~Ti (16\,GB), bfloat16 mixed precision, 8-bit AdamW~\citep{loshchilov2019decoupled}, learning rate $2 \times 10^{-4}$ with cosine decay, 700 epochs, batch size~32.
FlowResNet: 64 features, 16 blocks, 128-dim time embedding, 4 exits at blocks 3, 7, 11, 15.
EMA decay 0.9999.

\subsection{Main Results}
\label{sec:main_results}

\begin{table}[t]
\centering
\caption{Quantitative comparison on multi-organ histopathology $\times$4 SR (343 validation patches). Best in \textbf{bold}, second-best \underline{underlined}. SR3 GFLOPs reflect total cost of 40 forward passes (10 DDIM steps $\times$ 4 averaged samples).}
\label{tab:main_x4}
\resizebox{\columnwidth}{!}{
\begin{tabular}{lccccc}
\toprule
Method & Params & GFLOPs & PSNR$\uparrow$ & SSIM$\uparrow$ & LPIPS$\downarrow$ \\
\midrule
Bicubic & --- & --- & 29.24 & 0.8096 & 0.2495 \\
EDSR & 1.52M & 16.25 & 31.44 & 0.8316 & 0.2003 \\
SwinIR-light & 0.93M & 8.72 & 31.76 & 0.8494 & \underline{0.1953} \\
SwinIR-Medium & 11.90M & 107.11 & \textbf{31.84} & \underline{0.8790} & 0.1958 \\
SRFormer-light & 0.87M & 8.25 & 31.55 & 0.8718 & 0.1955 \\
SR3 & 2.94M & 4376 & 31.39 & 0.8672 & \textbf{0.1917} \\
\midrule
\ours{} (Exit~3) & 1.90M & 13.34 & \textbf{31.84} & \textbf{0.8797} & 0.1961 \\
\ours{} (Adaptive) & 1.90M & 8.92 & 31.72 & 0.8737 & 0.1967 \\
\bottomrule
\end{tabular}}
\end{table}

Table~\ref{tab:main_x4} compares \ours{} against baselines on 343 validation patches spanning breast, kidney, and lung tissue.
At full depth (Exit~3), \ours{} reaches 31.84\,dB PSNR and the highest SSIM (0.8797). SwinIR-Medium matches this PSNR (31.84\,dB) and attains slightly lower SSIM (0.8790), but requires 11.90M parameters and 107.11\,GFLOPs, compared with 1.90M parameters and 13.34\,GFLOPs for \ours{} Exit~3.
SRFormer-light, a recent ICCV~2023 transformer with permuted self-attention, achieves 31.55\,dB PSNR and 0.8718 SSIM with the fewest parameters (0.87M) and comparable GFLOPs to SwinIR-light.
With adaptive routing, \ours{} maintains 31.72\,dB at only 8.9\,GFLOPs, comparable to both SwinIR-light (8.72\,GFLOPs) and SRFormer-light (8.25\,GFLOPs), while achieving substantially higher structural similarity (SSIM $+$0.024 over SwinIR-light, $+$0.002 over SRFormer-light).
The 33\% compute savings versus full-depth processing come at a cost of only 0.12\,dB PSNR, demonstrating that the router effectively identifies images where full depth is unnecessary.
On LPIPS, SR3 remains best (0.1917); among the regression baselines, SwinIR-light (0.1953) and SwinIR-Medium (0.1958) are slightly ahead of \ours{} Exit~3 (0.1961), while \ours{} retains the strongest compute-quality tradeoff.

SR3 uses SDEdit-style initialization (noisy bicubic at $t{=}200$) with 10 DDIM steps and 4-sample averaging, totaling 40 forward passes at full HR resolution (4{,}376\,GFLOPs).
This yields competitive distortion metrics (31.39\,dB PSNR) and the best LPIPS (0.1917), consistent with diffusion models' strength in perceptual quality~\citep{saharia2022image}.
However, this comes at ${\sim}490{\times}$ the compute of \ours{} (Adaptive, 8.92\,GFLOPs), which achieves higher PSNR (+0.33\,dB) and SSIM (+0.007).
\ours{} thus achieves superior distortion quality at a fraction of SR3's compute cost, validating the efficiency of single-step flow matching with adaptive depth routing.
Moreover, SR3 operates at full HR resolution ($256{\times}256$, 6 channels), whereas \ours{} processes in pixel-unshuffled rearranged space ($64{\times}64$, 48 channels), yielding a substantially smaller memory footprint (see Table~\ref{tab:latency}).

All pairwise differences are assessed via paired Wilcoxon signed-rank tests over the 343 validation images.
\ours{} Exit~3 significantly outperforms EDSR, SRFormer-light, SR3, and SwinIR-light on both PSNR and SSIM ($p < 0.001$ in all cases).
Compared to SwinIR-Medium, \ours{} Exit~3 achieves significantly higher SSIM ($p < 0.01$) and statistically indistinguishable LPIPS ($p = 0.82$), while matching PSNR to two decimal places in Table~\ref{tab:main_x4} at $8{\times}$ fewer GFLOPs.
The adaptive quality cost ($-$0.12\,dB PSNR, $-$0.006 SSIM vs.\ full depth) is statistically significant ($p < 0.001$) but practically small, representing the price of a 33\% compute reduction; the corresponding measured latency reduction is 34\% (Section~\ref{sec:latency}).

\subsection{Inference Efficiency}
\label{sec:latency}

\begin{table}[t]
\centering
\caption{Inference efficiency on a single RTX~5070~Ti (16\,GB), bfloat16, batch~1, $64{\times}64$ LR input ($\times$4 SR). Latency averaged over 200 runs after 20 warmup passes. Best learned method per metric in \textbf{bold}.}
\label{tab:latency}
\small
\setlength{\tabcolsep}{3pt}
\resizebox{\columnwidth}{!}{%
\begin{tabular}{lrrrrc}
    \toprule
    Method & Params & GFLOPs & ms & img/s & VRAM \\
    \midrule
    Bicubic & --- & --- & 0.9 & 1{,}125 & 2 \\
    EDSR & 1.52M & 16.25 & 20.8 & 48 & 30 \\
    SwinIR-light & 0.93M & 8.72 & 60.1 & 17 & 99 \\
    SwinIR-Medium & 11.90M & 107.11 & 84.1 & 12 & 129 \\
    SRFormer-light & \textbf{0.87M} & 8.25 & 44.9 & 22 & 106 \\
    SR3 & 2.94M & 4{,}376 & 1{,}082.1 & 1 & 145 \\
    \midrule
    \ours{} (E0) & 1.90M & \textbf{3.10} & \textbf{2.4} & \textbf{418} & \textbf{22} \\
    \ours{} (E1) & 1.90M & 6.07 & 4.7 & 211 & 43 \\
    \ours{} (E2) & 1.90M & 9.39 & 8.2 & 122 & 47 \\
    \ours{} (E3) & 1.90M & 13.34 & 12.4 & 80 & 49 \\
    \ours{} (Adp) & 1.90M & 8.92 & 8.2 & 122 & 46 \\
    \bottomrule
\end{tabular}}
\end{table}

Table~\ref{tab:latency} reports wall-clock latency and peak GPU memory measured on a single RTX~5070~Ti with bfloat16 mixed precision and batch size~1.
GFLOPs translate to proportional latency gains: \ours{} Exit~0 (2.4\,ms) is the fastest learned method, and Exit~3 (12.4\,ms) is faster than both SwinIR-light (60.1\,ms) and SwinIR-Medium (84.1\,ms).
Peak GPU memory usage (VRAM) follows the same trend: \ours{} uses 22 to 49\,MB across exits versus 99\,MB for SwinIR-light, 129\,MB for SwinIR-Medium, and 145\,MB for SR3, directly reflecting the benefit of operating at $64{\times}64$ rearranged resolution rather than $256{\times}256$ HR.
SR3's 40 forward passes incur $1{,}082$\,ms latency ($87{\times}$ slower than \ours{} E3), confirming that GFLOPs savings translate to real wall-clock speedups.
Notably, \ours{} E3 is faster than SwinIR-light despite having more GFLOPs (13.34 vs.\ 8.72), and is $6.8{\times}$ faster than SwinIR-Medium while using roughly $8{\times}$ fewer GFLOPs (13.34 vs.\ 107.11). The convolution-dominated rearranged-space formulation is memory-bandwidth efficient, whereas SwinIR's window attention incurs overhead from reshaping, masking, and softmax operations that underutilize GPU compute at small batch sizes.
Adaptive routing achieves the same 34\% latency reduction predicted by GFLOPs: \ours{} Adaptive processes each tile in 8.2\,ms versus 12.4\,ms for Exit~3, with the exit classifier routing each tile to its predicted optimal depth at negligible overhead.
In a clinical digital pathology pipeline processing hundreds of tiles per whole-slide image, these per-tile latency differences compound: at Exit~3, \ours{} can process a 300-tile slide in ${\sim}$3.7\,s versus ${\sim}$325\,s for SR3, a difference that determines whether real-time SR is feasible during diagnostic review.

\subsection{Per-Exit Analysis}
\label{sec:per_exit}

\begin{table}[t]
\centering
\caption{Per-exit quality and compute breakdown for \ours{} on $\times$4 SR. Hybrid shows the number of HybridFiLMBlocks processed at each exit.}
\label{tab:exits}
\small
\setlength{\tabcolsep}{4pt}
\resizebox{\columnwidth}{!}{%
\begin{tabular}{lcccccc}
\toprule
Exit & Blocks & Hybrid & GFLOPs & PSNR$\uparrow$ & SSIM$\uparrow$ & LPIPS$\downarrow$ \\
\midrule
E0 & 0--3 & 0 & 3.10 & 31.17 & 0.8625 & 0.2003 \\
E1 & 0--7 & 1 & 6.07 & 31.60 & 0.8717 & 0.1969 \\
E2 & 0--11 & 3 & 9.39 & 31.72 & 0.8740 & 0.1971 \\
E3 & 0--15 & 6 & 13.34 & 31.84 & 0.8797 & 0.1961 \\
\midrule
Adaptive & --- & --- & 8.92 & 31.72 & 0.8737 & 0.1967 \\
\bottomrule
\end{tabular}}
\end{table}

Table~\ref{tab:exits} shows the per-exit quality and compute breakdown.
Quality gains exhibit diminishing returns with depth: the largest gain occurs at E0$\to$E1 ($+$0.43\,dB), while E1$\to$E2 and E2$\to$E3 each contribute only $+$0.12\,dB.
Meanwhile, compute increases super-linearly due to the concentration of hybrid attention blocks in later exits: adding the final 4 blocks (E2$\to$E3) costs 3.95\,GFLOPs (42\% of the full cost) for only 0.12\,dB improvement.
This asymmetry between quality saturation and compute growth is precisely what motivates adaptive routing.

Notably, Exit~0 alone (3.1\,GFLOPs, pure convolution) already exceeds bicubic by $+$1.93\,dB and approaches EDSR quality (31.17 vs.\ 31.44\,dB) at $5.2\times$ less compute, demonstrating that the rearranged-space formulation enables strong baselines even at minimal depth.
The adaptive strategy matches Exit~2 quality (31.72\,dB) at 5\% less compute (8.92 vs.\ 9.39\,GFLOPs), confirming that the router effectively identifies images that do not benefit from the expensive attention blocks in the final segment.
This graduated design (pure convolution in the first segment, progressively more attention in later segments) means that the marginal cost of each quality increment increases, giving the router a natural trade-off surface to exploit.

\subsection{Ablation Study}
\label{sec:ablation}

\begin{table}[ht!]
\centering
\caption{Ablation study on $\times$4 SR. Each row removes one component from the full \ours{} model. $\Delta$PSNR shows the drop from the full model. Ablations trained for 300 epochs; full model for 700 epochs. Best per metric in \textbf{bold}.}
\label{tab:ablation}
\small
\setlength{\tabcolsep}{4pt}
\resizebox{\columnwidth}{!}{%
\begin{tabular}{lcccc}
\toprule
Variant & PSNR$\uparrow$ & SSIM$\uparrow$ & LPIPS$\downarrow$ & $\Delta$PSNR \\
\midrule
Full \ours{} & \textbf{31.72} & 0.8737 & 0.1967 & -- \\
w/o early exit & 31.65 & \textbf{0.8742} & \textbf{0.1959} & $-0.07$ \\
w/o consistency & 31.55 & 0.8717 & 0.1975 & $-0.17$ \\
w/o $t{=}0$ mixing & 30.17 & 0.8414 & 0.2175 & $-1.54$ \\
w/o SSIM loss & 31.41 & 0.8680 & 0.1978 & $-0.30$ \\
Fewer blocks (8) & 31.41 & 0.8677 & 0.1968 & $-0.31$ \\
\bottomrule
\end{tabular}}
\end{table}

Table~\ref{tab:ablation} ablates key components of the full \ours{} model.
The most critical factor is $t{=}0$ mixing ($-$1.54\,dB without it): dedicating half of training to exact $t{=}0$ samples is essential for single-step inference quality, as the model otherwise lacks direct supervision at the inference timestep.
Without $t{=}0$ mixing, the model is trained predominantly on intermediate timesteps ($t > 0$) where the input $\mathbf{x}_t$ already contains partial information about the target; at inference ($t{=}0$), the model faces a distributional shift since the input is pure bicubic with no target content, leading to the large quality drop.
This finding has practical significance for clinical deployment: since inference always evaluates the model at $t{=}0$, the training distribution must be aligned with the deployment distribution, making $t{=}0$ mixing not merely a training heuristic but a necessary condition for reliable single-step flow matching SR.

SSIM loss ($-$0.30\,dB) provides structural supervision that complements the per-pixel L1 velocity loss: L1 alone tends to produce slightly blurred reconstructions, whereas SSIM penalizes degradation of local luminance, contrast, and structural patterns.
Fewer blocks ($-$0.31\,dB) demonstrates that the full 16-block depth is necessary; halving to 8 blocks removes all 6 hybrid attention blocks, eliminating the global receptive field needed for long-range structural coherence in complex tissue.

Consistency loss contributes $+$0.17\,dB by encouraging the model to predict the same target image from different timesteps, improving prediction stability.
This regularization encourages a smooth velocity field: without it, the model may overfit to specific timestep ranges, producing inconsistent $\mathbf{x}_0$ predictions that degrade average quality.

Removing the early exit mechanism (forcing all images through full depth) causes only $-$0.07\,dB PSNR change with comparable SSIM and LPIPS, confirming that the multi-exit training strategy introduces no quality penalty at the final exit; the early exits act as auxiliary supervision without degrading the main output.
The ``w/o early exit'' variant achieves marginally better SSIM (0.8742 vs.\ 0.8737) and LPIPS (0.1959 vs.\ 0.1967), suggesting that auxiliary exit supervision introduces a slight tension between exit-level and final-output objectives, though the magnitude is negligible ($<$0.1\%).
The interplay between $t{=}0$ mixing and consistency loss is noteworthy: $t{=}0$ mixing provides direct single-step supervision, while consistency loss regularizes the velocity field at intermediate timesteps, yielding a model that is both accurate at inference time and robust to timestep variation during training.

\subsection{Compute-Quality Pareto Frontier}
\label{sec:pareto}

\begin{figure}[!htbp]
\centering
\includegraphics[width=0.95\linewidth]{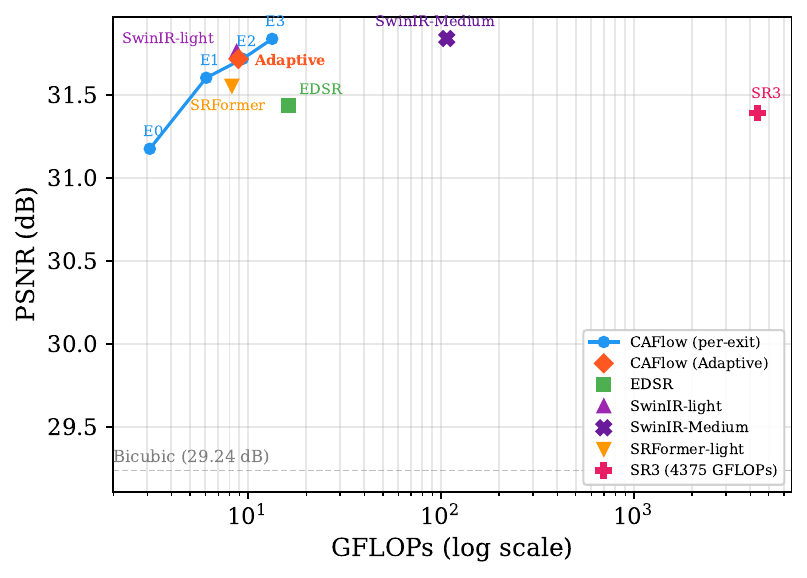}
\caption{Compute-quality Pareto frontier. \ours{} at different operating points (Exit~0 through Exit~3 and Adaptive) compared against fixed baselines including SwinIR-Medium. The adaptive strategy achieves near-full-depth quality at reduced compute. SwinIR-Medium reaches similar PSNR to \ours{} Exit~3 only at substantially higher compute, while the \ours{} frontier remains stronger in the practical 3.1 to 13.3\,GFLOPs regime.}
\label{fig:pareto}
\end{figure}

Figure~\ref{fig:pareto} plots PSNR against GFLOPs for \ours{} at each exit and the adaptive operating point, revealing a smooth compute-quality tradeoff.
The key finding is that \ours{} provides a much stronger compute-quality tradeoff than the fixed baselines: even the cheapest Exit~0 (31.17\,dB, 3.1\,GFLOPs) approaches EDSR quality (31.44\,dB, 16.25\,GFLOPs) at $5.2{\times}$ less compute, while Exit~1 (31.60\,dB, 6.1\,GFLOPs) surpasses both EDSR and SRFormer-light (31.55\,dB, 8.25\,GFLOPs) at less compute.
The adaptive operating point (31.72\,dB, 8.9\,GFLOPs) achieves quality on par with SwinIR-light (31.76\,dB, 8.72\,GFLOPs) and surpasses SRFormer-light at comparable compute, but with substantially higher SSIM (0.8737 vs.\ 0.8494 and 0.8718, respectively).
SwinIR-Medium reaches comparable full-depth quality (31.84\,dB) only at 107.11\,GFLOPs, placing it far to the right of the practical operating region covered by \ours{}.
This demonstrates that the graduated backbone design, with cheap convolutions for easy images and expensive attention for hard ones, provides a more efficient compute-quality tradeoff than fixed-architecture baselines.
The Pareto curve also shows that no single fixed-compute baseline lies on the \ours{} frontier in the low- to moderate-compute regime: EDSR (16.25\,GFLOPs) is Pareto-dominated by Exit~1 at $2.7\times$ less compute, SRFormer-light (8.25\,GFLOPs) is surpassed by Exit~1, and SwinIR-light (8.72\,GFLOPs) matches the adaptive point in PSNR but at lower SSIM, confirming the value of adaptive depth allocation over fixed-architecture design.

\subsection{Exit Distribution}
\label{sec:exit_dist}

\begin{figure}[!htbp]
\centering
\includegraphics[width=0.95\linewidth]{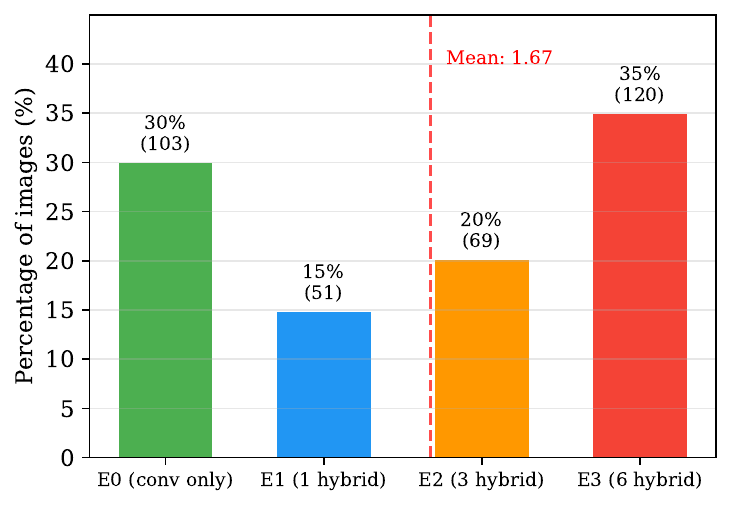}
\caption{Exit distribution across the 343 validation images under adaptive routing. The classifier routes a mix of images to each exit, with many easy images handled at E0/E1 and harder images pushed to E2/E3.}
\label{fig:exit_histogram}
\end{figure}

Figure~\ref{fig:exit_histogram} shows the distribution of predicted exits across the 343 validation images.
The classifier learns a non-trivial routing policy: it routes a substantial fraction of images to early exits (E0/E1), where cheap convolution-only processing suffices, while reserving deeper exits (E2/E3) with expensive attention blocks for images with dense cellular structures, complex nuclear morphology, or fine-grained texture that benefit from global receptive field.
This distribution yields the aggregate 8.9\,GFLOPs compute cost, a 33\% reduction versus processing all images at full depth (13.3\,GFLOPs).
Importantly, the router achieves this saving while sacrificing only 0.12\,dB PSNR (31.72 vs.\ 31.84\,dB), confirming that early exits do not degrade quality for the images routed to them.
Consistent with this, router predictions fall within one exit of the oracle assignment for 97.1\% of validation images, indicating that the learned policy is near-oracle even when exact exit labels differ.
The classifier's routing decisions correlate with image content: homogeneous stromal and adipose patches are predominantly routed to E0/E1, while patches containing dense glandular epithelium, heterogeneous staining, or fine-grained nuclear detail are pushed to E2/E3.

\subsection{Whole-Slide Tissue Analysis}
\label{sec:wsi}

\begin{figure}[!htbp]
\centering
\includegraphics[width=\columnwidth]{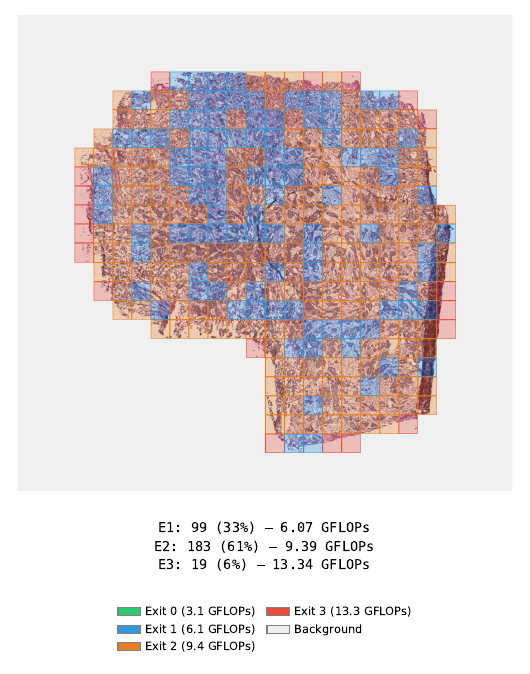}
\caption{Spatial exit assignment on a held-out TCGA-BRCA frozen tissue slide (TCGA-A8-A08C, not used during training) tiled at 40$\times$ into 1024$\times$1024 patches. After tissue detection (Otsu thresholding), 301 tissue tiles are retained and colored by their router-assigned exit: blue (E1, 6.07\,GFLOPs), orange (E2, 9.39\,GFLOPs), or red (E3, 13.34\,GFLOPs). The router assigns 33\% to E1 and 61\% to E2, with only 6\% routed to E3, averaging 8.55\,GFLOPs (a 36\% reduction versus full-depth processing).}
\label{fig:wsi}
\end{figure}

To demonstrate clinical applicability, we tile a held-out TCGA-BRCA frozen tissue whole-slide image (TCGA-A8-A08C, not present in the training data) at native 40$\times$ magnification into $1024{\times}1024$ patches and apply \ours{}'s exit routing to 301 tissue tiles identified via Otsu thresholding (Figure~\ref{fig:wsi}).
The resulting spatial exit map reveals spatially coherent routing decisions: the classifier routes 33\% of tiles to E1 (6.07\,GFLOPs) and 61\% to E2 (9.39\,GFLOPs), with only 6\% assigned to E3 (13.34\,GFLOPs), yielding an average cost of 8.55\,GFLOPs per tile, a 36\% reduction versus full-depth processing.
Tiles containing homogeneous tissue content are consistently routed to the shallower E1, while tiles at the boundary between tissue and background, where mixed content creates harder reconstruction targets, are assigned to deeper exits (E2, E3).
Notably, this spatial correspondence emerges purely from reconstruction quality signals: the classifier receives no tissue-type labels, no segmentation maps, and no spatial information, yet it learns a routing policy that aligns with tissue complexity as perceived by a pathologist.
This result suggests that adaptive SR routing could serve as a lightweight tissue complexity indicator in whole-slide analysis pipelines.
From a deployment perspective, the 36\% compute reduction at the slide level translates directly to lower inference cost and faster turnaround in high-throughput digital pathology workflows, where thousands of slides may require processing daily.

\FloatBarrier
\subsection{Qualitative Comparison}
\label{sec:qualitative}

\begin{figure*}[!htbp]
\centering
\includegraphics[width=0.99\textwidth]{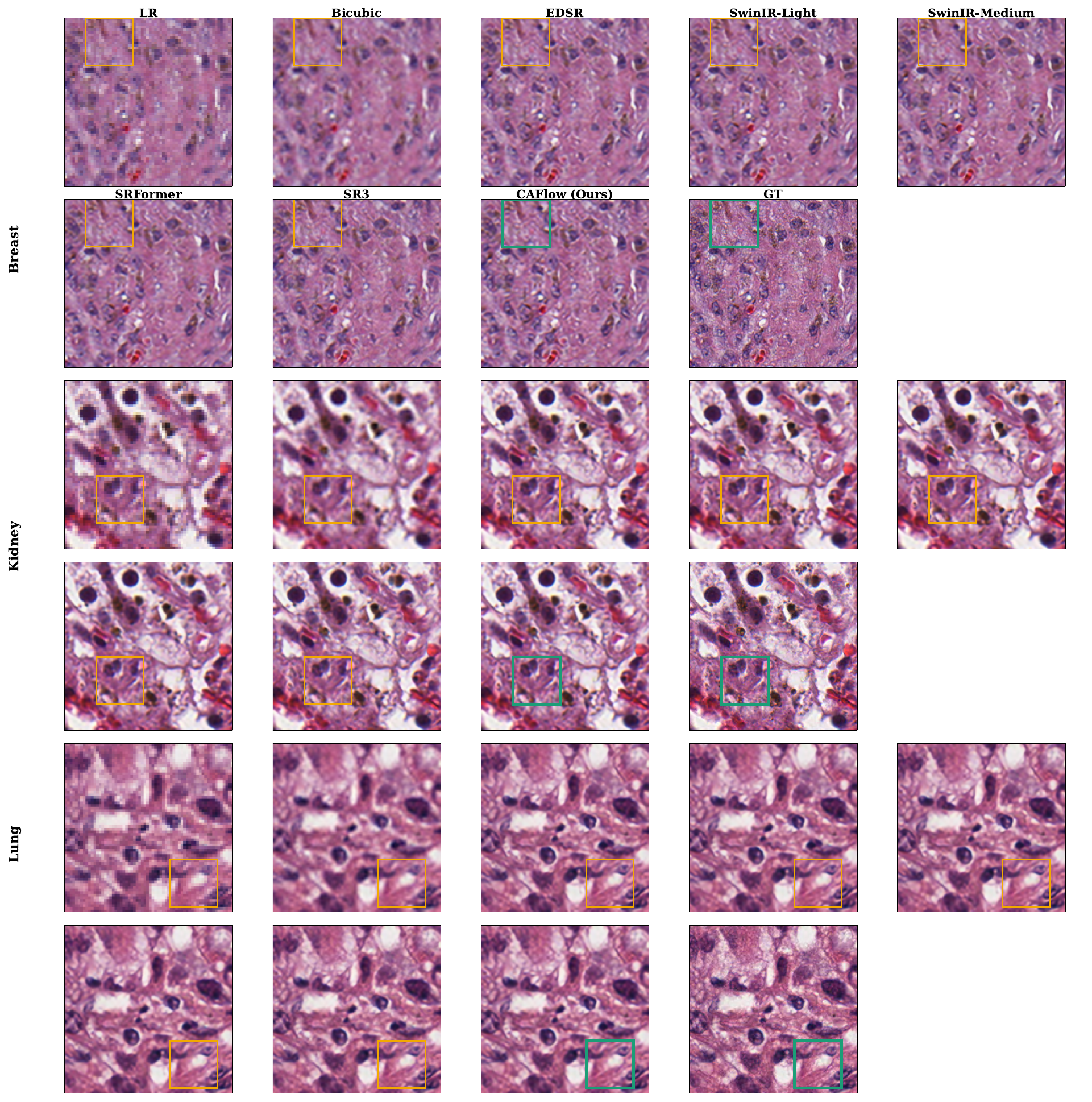}
\caption{Qualitative comparison on $\times$4 histopathology SR across breast, kidney, and lung tissue.
All examples are taken from the deterministic validation split, using organ-specific patches with highlighted ROIs.
Colored boxes mark structures where CAFlow preserves clearer or competitive nuclear separation, lumen definition, or alveolar/stromal continuity than competing methods. Across the highlighted ROIs, CAFlow matches SwinIR-Medium in mean PSNR within 0.03\,dB (25.61 vs.\ 25.63\,dB) while achieving higher mean SSIM (0.760 vs.\ 0.751). Although SwinIR-Medium and SR3 can be visually competitive, they are roughly one to two orders of magnitude heavier in compute and memory.}
\label{fig:qualitative}
\end{figure*}

Figure~\ref{fig:qualitative} shows visual comparisons across three tissue types (breast, kidney, lung).
On easier stromal and tubular patches (breast, kidney), all learned methods substantially outperform bicubic interpolation, and \ours{} at full depth achieves PSNR comparable to SwinIR-light despite using a fundamentally different generative framework.
On the most challenging patch (lung, 23.71\,dB bicubic), the dense alveolar structures push all methods below 27\,dB, yet \ours{} (26.99\,dB) and SwinIR-light (27.00\,dB) are virtually tied, demonstrating that flow matching achieves competitive perceptual quality without multi-step diffusion.
EDSR tends to over-smooth fine cellular boundaries, particularly in the lung tissue where overlapping alveolar walls require high-frequency detail preservation; the attention blocks in \ours{}'s deeper exits help capture these long-range structural dependencies.
SR3 achieves similar PSNR but requires 40 forward passes at HR resolution (4{,}376\,GFLOPs), whereas \ours{} uses a single forward pass at LR resolution (13.3\,GFLOPs for full depth, or 8.9\,GFLOPs with adaptive routing). SwinIR-Medium also reaches similar full-depth quality, but at 107.11\,GFLOPs, making \ours{} substantially lighter than both the heavy transformer and multi-step diffusion alternatives in this qualitative comparison.
Importantly, the single-step formulation does not sacrifice perceptual sharpness: across all three tissue types, \ours{} produces edges and textures visually comparable to multi-step SR3, suggesting that flow matching's direct velocity prediction captures high-frequency details without iterative refinement.

\subsection{Per-Organ Analysis}
\label{sec:per_organ}

\begin{table*}[t]
\centering
\caption{Per-organ quantitative comparison on multi-organ histopathology $\times$4 SR. Values are mean $\pm$ std across validation patches per organ. Best in \textbf{bold}, second-best \underline{underlined}.}
\label{tab:per_organ}
\small
\setlength{\tabcolsep}{3pt}
\resizebox{\textwidth}{!}{%
\begin{tabular}{lccccccccc}
\toprule
& \multicolumn{3}{c}{Breast} & \multicolumn{3}{c}{Kidney} & \multicolumn{3}{c}{Lung} \\
\cmidrule(lr){2-4} \cmidrule(lr){5-7} \cmidrule(lr){8-10}
Method & PSNR$\uparrow$ & SSIM$\uparrow$ & LPIPS$\downarrow$ & PSNR$\uparrow$ & SSIM$\uparrow$ & LPIPS$\downarrow$ & PSNR$\uparrow$ & SSIM$\uparrow$ & LPIPS$\downarrow$ \\
\midrule
Bicubic & 30.64\tiny{$\pm$4.41} & 0.8331\tiny{$\pm$0.0928} & 0.2293\tiny{$\pm$0.0523} & 27.35\tiny{$\pm$5.76} & 0.7777\tiny{$\pm$0.0957} & 0.2765\tiny{$\pm$0.0686} & 27.80\tiny{$\pm$3.92} & 0.7932\tiny{$\pm$0.0867} & 0.2639\tiny{$\pm$0.0527} \\
EDSR & 32.97\tiny{$\pm$4.42} & 0.8834\tiny{$\pm$0.0727} & 0.1837\tiny{$\pm$0.0506} & 29.57\tiny{$\pm$5.30} & 0.8499\tiny{$\pm$0.0734} & 0.2204\tiny{$\pm$0.0642} & 30.13\tiny{$\pm$4.06} & 0.8574\tiny{$\pm$0.0659} & 0.2138\tiny{$\pm$0.0525} \\
SwinIR-light & 33.23\tiny{$\pm$4.40} & 0.8883\tiny{$\pm$0.0701} & 0.1808\tiny{$\pm$0.0496} & 30.05\tiny{$\pm$5.37} & 0.8611\tiny{$\pm$0.0675} & 0.2148\tiny{$\pm$0.0586} & 30.46\tiny{$\pm$3.97} & 0.8650\tiny{$\pm$0.0627} & 0.2099\tiny{$\pm$0.0511} \\
SwinIR-Medium & \textbf{33.39\tiny{$\pm$4.40}} & \underline{0.8913\tiny{$\pm$0.0687}} & \underline{0.1789\tiny{$\pm$0.0497}} & \textbf{30.27\tiny{$\pm$5.33}} & \textbf{0.8665\tiny{$\pm$0.0647}} & 0.2165\tiny{$\pm$0.0579} & \textbf{30.66\tiny{$\pm$4.04}} & \underline{0.8691\tiny{$\pm$0.0611}} & 0.2099\tiny{$\pm$0.0513} \\
SRFormer-light & 33.04\tiny{$\pm$4.36} & 0.8852\tiny{$\pm$0.0714} & 0.1812\tiny{$\pm$0.0494} & 29.84\tiny{$\pm$5.39} & 0.8559\tiny{$\pm$0.0706} & \underline{0.2125\tiny{$\pm$0.0605}} & 30.26\tiny{$\pm$3.92} & 0.8611\tiny{$\pm$0.0639} & \underline{0.2088\tiny{$\pm$0.0510}} \\
SR3 & 32.95\tiny{$\pm$4.50} & 0.8824\tiny{$\pm$0.0745} & \textbf{0.1772\tiny{$\pm$0.0462}} & 29.63\tiny{$\pm$5.46} & 0.8498\tiny{$\pm$0.0737} & \textbf{0.2074\tiny{$\pm$0.0539}} & 30.09\tiny{$\pm$4.14} & 0.8554\tiny{$\pm$0.0678} & \textbf{0.2033\tiny{$\pm$0.0475}} \\
\midrule
\ours{} (E3) & \underline{33.29\tiny{$\pm$4.52}} & \textbf{0.8921\tiny{$\pm$0.0670}} & 0.1799\tiny{$\pm$0.0509} & \underline{30.08\tiny{$\pm$5.48}} & \underline{0.8658\tiny{$\pm$0.0646}} & 0.2140\tiny{$\pm$0.0574} & \underline{30.51\tiny{$\pm$4.19}} & \textbf{0.8698\tiny{$\pm$0.0598}} & 0.2096\tiny{$\pm$0.0503} \\
\ours{} (Adaptive) & 33.15\tiny{$\pm$4.52} & 0.8864\tiny{$\pm$0.0720} & 0.1807\tiny{$\pm$0.0511} & 29.97\tiny{$\pm$5.46} & 0.8601\tiny{$\pm$0.0669} & 0.2141\tiny{$\pm$0.0582} & 30.39\tiny{$\pm$4.20} & 0.8630\tiny{$\pm$0.0642} & 0.2102\tiny{$\pm$0.0504} \\
\bottomrule
\end{tabular}}
\end{table*}

Table~\ref{tab:per_organ} reports per-organ metrics with standard deviations across validation patches for breast, kidney, and lung tissue.
Performance varies across organs: breast tissue, with relatively homogeneous stromal patterns, achieves the highest PSNR across all methods; kidney tissue, containing fine glomerular structures and tubular boundaries, presents intermediate difficulty; and lung tissue, with dense overlapping alveolar walls and heterogeneous cellular density, is consistently the most challenging.
These organ-specific differences are reflected in the standard deviations: lung patches exhibit wider variance in all metrics, reflecting the greater diversity of tissue structures within the organ.

The adaptive router's exit distribution also varies by organ (Section~\ref{sec:exit_dist}): breast patches are more frequently routed to early exits (E0/E1), consistent with their lower reconstruction difficulty, while lung patches tend to require deeper exits (E2/E3) due to the fine-grained alveolar detail that benefits from the attention layers.
Kidney presents a middle ground, with the router distributing images across all exits roughly uniformly.
These per-organ routing patterns emerge without any tissue-type supervision, confirming that the quality-aware exit classifier captures intrinsic differences in reconstruction difficulty across anatomical structures.

\subsection{Downstream Task Validation}
\label{sec:downstream}

\begin{table}[t]
\centering
\caption{Downstream nuclei detection on SR outputs. StarDist~\citep{schmidt2018stardist} ``2D\_versatile\_he'' detects nuclei on each method's SR output; metrics compare against detections on HR ground truth (IoU threshold = 0.5). All learned SR methods significantly outperform bicubic interpolation; differences among learned methods are not statistically significant (paired $t$-test, $p > 0.05$), indicating that downstream clinical utility has saturated while computational cost varies widely (Table~\ref{tab:latency}).}
\label{tab:downstream}
\small
\setlength{\tabcolsep}{3pt}
\resizebox{\columnwidth}{!}{%
\begin{tabular}{lccccc}
\toprule
Method & Precision$\uparrow$ & Recall$\uparrow$ & Detection F1$\uparrow$ & Count $r$$\uparrow$ & Mean IoU$\uparrow$ \\
\midrule
Bicubic & 0.863\tiny{$\pm$0.152} & 0.886\tiny{$\pm$0.132} & 0.869\tiny{$\pm$0.132} & 0.9799 & 0.847\tiny{$\pm$0.080} \\
EDSR & 0.898\tiny{$\pm$0.127} & 0.926\tiny{$\pm$0.095} & 0.905\tiny{$\pm$0.109} & 0.9878 & 0.879\tiny{$\pm$0.065} \\
SwinIR-light & 0.906\tiny{$\pm$0.122} & 0.928\tiny{$\pm$0.099} & 0.911\tiny{$\pm$0.108} & 0.9886 & 0.883\tiny{$\pm$0.064} \\
SwinIR-Medium & 0.895\tiny{$\pm$0.126} & 0.936\tiny{$\pm$0.102} & 0.909\tiny{$\pm$0.111} & 0.9868 & 0.884\tiny{$\pm$0.068} \\
SRFormer-light & 0.913\tiny{$\pm$0.111} & 0.916\tiny{$\pm$0.111} & 0.910\tiny{$\pm$0.101} & 0.9895 & 0.885\tiny{$\pm$0.042} \\
SR3 & 0.901\tiny{$\pm$0.122} & 0.924\tiny{$\pm$0.100} & 0.906\tiny{$\pm$0.106} & 0.9866 & 0.880\tiny{$\pm$0.065} \\ \midrule
\ours{} (E3) & 0.902\tiny{$\pm$0.130} & 0.925\tiny{$\pm$0.101} & 0.906\tiny{$\pm$0.117} & 0.9888 & 0.881\tiny{$\pm$0.081} \\
\ours{} (Adaptive) & 0.905\tiny{$\pm$0.122} & 0.925\tiny{$\pm$0.100} & 0.909\tiny{$\pm$0.107} & 0.9873 & 0.882\tiny{$\pm$0.066} \\
\bottomrule
\end{tabular}}
\end{table}

A critical question for clinical deployment of SR in pathology is whether improved pixel-level metrics translate to preserved diagnostic information.
To evaluate this, we apply StarDist~\citep{schmidt2018stardist}, a widely-used nuclei detection model trained on Hematoxylin and Eosin (H\&E) histopathology (``2D\_versatile\_he''), to SR outputs from all methods and compare detected nuclei against detections on HR ground truth images.
We match SR detections to HR detections using Intersection over Union (IoU) $>$ 0.5 and report detection F1, Pearson count correlation, and mean IoU of matched nuclei (Table~\ref{tab:downstream}).

All learned SR methods significantly outperform bicubic interpolation (F1: 0.87 to 0.90/0.91), confirming that super-resolution preserves diagnostically relevant nuclear structures.
Crucially, differences among learned methods are not statistically significant (paired $t$-test, $p > 0.05$ for all pairwise comparisons), indicating that downstream nuclei detection has saturated: further pixel-metric improvements do not yield measurably better detection.
This saturation makes computational efficiency an important differentiator. \ours{} (Adaptive) achieves the same downstream utility as SwinIR-light ($\text{F1} = 0.909$ vs.\ $0.911$, $p = 0.84$) and matches SwinIR-Medium on detection F1 ($0.909$ vs.\ $0.909$), while requiring far less compute than the larger transformer baseline (Table~\ref{tab:latency}).
All methods achieve high count correlations ($r > 0.987$), with \ours{} (Exit~3) at $r = 0.989$, confirming that the rearranged-space formulation effectively preserves accurate nuclei counts.
These results validate \ours{} not merely as a visualization enhancement but as a preprocessing step for automated histopathology analysis pipelines, where efficiency directly impacts whole-slide throughput.

\subsection{Cross-Organ Generalization}
\label{sec:held_out}

A key concern for clinical deployment is whether SR models generalize to tissue types unseen during training.
To evaluate this, we test all methods on colon tissue (TCGA-COAD, 785 patches), an organ entirely absent from the training set (breast, kidney, lung).
Colon tissue is histologically distinct, featuring mucosal crypts, goblet cells, and lamina propria that differ substantially from the glandular, tubular, and alveolar structures in the training organs.

\begin{table}[t]
\centering
\caption{Zero-shot cross-organ generalization on held-out colon tissue ($\times$4 SR). Models trained on breast/kidney/lung, evaluated on entirely unseen colon organ. Best in \textbf{bold}, second-best \underline{underlined}.}
\label{tab:held_out}
\small
\resizebox{\columnwidth}{!}{%
\begin{tabular}{lccc}
\toprule
Method & PSNR$\uparrow$ & SSIM$\uparrow$ & LPIPS$\downarrow$ \\
\midrule
Bicubic & 29.58\tiny{$\pm$4.72} & 0.8104\tiny{$\pm$0.0815} & 0.2469\tiny{$\pm$0.0595} \\
EDSR & 31.49\tiny{$\pm$4.44} & 0.8620\tiny{$\pm$0.0630} & 0.2030\tiny{$\pm$0.0579} \\
SwinIR-light & 31.75\tiny{$\pm$4.34} & 0.8678\tiny{$\pm$0.0610} & 0.1994\tiny{$\pm$0.0564} \\
SwinIR-Medium & \textbf{31.90\tiny{$\pm$4.40}} & \underline{0.8709\tiny{$\pm$0.0602}} & 0.1991\tiny{$\pm$0.0574} \\
SRFormer-light & 31.60\tiny{$\pm$4.30} & 0.8650\tiny{$\pm$0.0617} & \underline{0.1987\tiny{$\pm$0.0557}} \\
SR3 & 31.42\tiny{$\pm$4.54} & 0.8592\tiny{$\pm$0.0655} & \textbf{0.1931\tiny{$\pm$0.0536}} \\ \midrule
\ours{} (E3) & \underline{31.82\tiny{$\pm$4.65}} & \textbf{0.8719\tiny{$\pm$0.0591}} & 0.1993\tiny{$\pm$0.0576} \\
\ours{} (Adaptive) & 31.71\tiny{$\pm$4.63} & 0.8659\tiny{$\pm$0.0625} & 0.1996\tiny{$\pm$0.0571} \\
\bottomrule
\end{tabular}}
\end{table}

Table~\ref{tab:held_out} shows that \ours{} generalizes well to unseen colon tissue: Exit~3 achieves 31.82\,dB PSNR and the best SSIM (0.8719), while SwinIR-Medium attains slightly higher PSNR (31.90\,dB) at much higher compute.
Paired Wilcoxon tests over the 785 colon images confirm that \ours{} Exit~3 significantly outperforms EDSR and SwinIR-light ($p < 0.01$), and remains within 0.09\,dB of SwinIR-Medium ($p < 0.001$) at $8{\times}$ lower compute.
The performance gap between training organs (Table~\ref{tab:main_x4}) and held-out colon is minimal for \ours{} ($-$0.02\,dB PSNR), indicating that the learned representations capture general image reconstruction principles rather than organ-specific features.
The adaptive router distributes colon images primarily across intermediate exits (46\% E1, 47\% E2), treating colon as medium difficulty, a sensible assignment given that colon tissue has moderate structural complexity compared to the heterogeneous lung patches that frequently require deeper exits.
SR3 again achieves the best LPIPS (0.1932) on the held-out organ, consistent with its perceptual quality strength observed on training organs.

\subsection{Scaling to $\times$8 Super-Resolution}
\label{sec:x8}

\begin{table}[t]
\centering
\caption{Quantitative comparison on multi-organ histopathology $\times$8 SR. Best in \textbf{bold}, second-best \underline{underlined}.}
\label{tab:x8}
\small
\resizebox{\columnwidth}{!}{%
\begin{tabular}{lccc}
\toprule
Method & PSNR$\uparrow$ & SSIM$\uparrow$ & LPIPS$\downarrow$ \\
\midrule
Bicubic & 24.13\tiny{$\pm$4.58} & 0.6149\tiny{$\pm$0.1538} & 0.4303\tiny{$\pm$0.0733} \\
EDSR & 25.43\tiny{$\pm$4.45} & 0.6684\tiny{$\pm$0.1410} & 0.3800\tiny{$\pm$0.0697} \\
SwinIR-light & 25.30\tiny{$\pm$4.46} & 0.6616\tiny{$\pm$0.1399} & 0.3785\tiny{$\pm$0.0680} \\
SwinIR-Medium & \textbf{26.52\tiny{$\pm$4.61}} & \textbf{0.7105\tiny{$\pm$0.1316}} & 0.3555\tiny{$\pm$0.0711} \\
SRFormer-light & 25.23\tiny{$\pm$4.44} & 0.6584\tiny{$\pm$0.1406} & 0.3820\tiny{$\pm$0.0678} \\
SR3 & 25.57\tiny{$\pm$4.61} & 0.6728\tiny{$\pm$0.1430} & \textbf{0.3504\tiny{$\pm$0.0659}} \\ \midrule
\ours{} (E3) & \underline{26.21\tiny{$\pm$4.67}} & \underline{0.7047\tiny{$\pm$0.1321}} & 0.3557\tiny{$\pm$0.0701} \\
\ours{} (Adaptive) & 26.14\tiny{$\pm$4.63} & 0.6946\tiny{$\pm$0.1368} & \underline{0.3532\tiny{$\pm$0.0704}} \\
\bottomrule
\end{tabular}}
\end{table}

\begin{figure*}[!htbp]
\centering
\includegraphics[width=0.99\textwidth]{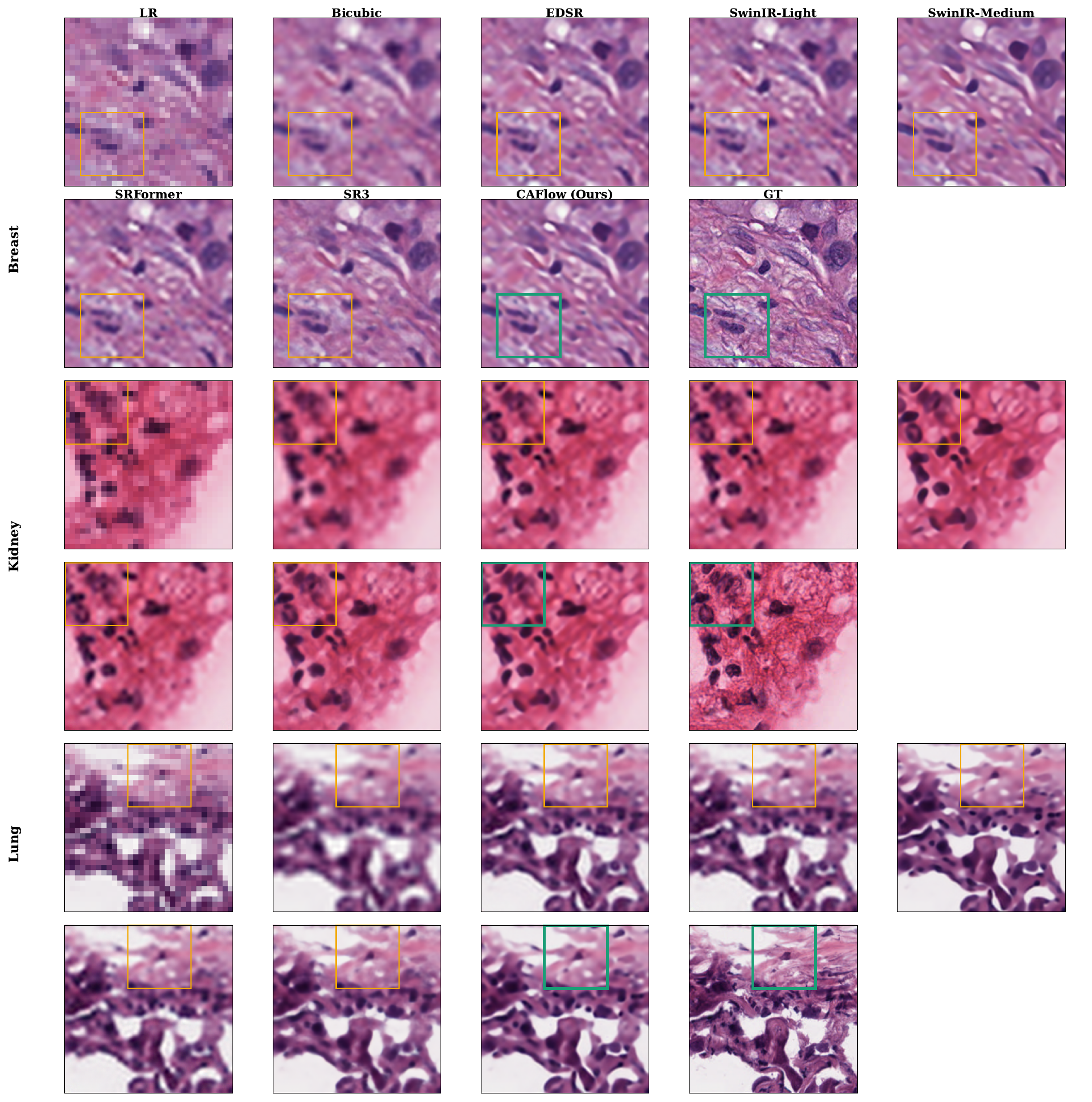}
\caption{Qualitative comparison on $\times$8 histopathology SR across breast, kidney, and lung tissue.
All examples are taken from the deterministic validation split, using organ-specific patches with highlighted ROIs.
Colored boxes mark structures where CAFlow preserves clearer or competitive nuclear separation and boundary continuity than competing methods at this harder scale. Across the highlighted ROIs, CAFlow matches SwinIR-Medium in mean PSNR within 0.03\,dB (23.55 vs.\ 23.57\,dB) while achieving higher mean SSIM (0.624 vs.\ 0.612). Although SwinIR-Medium and SR3 can be visually competitive, they are roughly one to two orders of magnitude heavier in compute and memory.}
\label{fig:qualitative_x8}
\end{figure*}

To evaluate whether \ours{}'s advantages extend to more challenging upscaling factors, we evaluated all methods at $\times$8 scale ($32{\times}32$ LR $\to$ $256{\times}256$ HR). All baselines were trained from scratch under the same x8 protocol; due to its substantially larger cost, SwinIR-Medium was trained for 100 epochs.
Table~\ref{tab:x8} presents the quantitative results and Figure~\ref{fig:qualitative_x8} shows representative visual comparisons.

At $\times$8, \ours{} is especially strong in the practical compute regime. Exit~3 achieves 26.21\,dB PSNR and 0.7047 SSIM, outperforming all comparable-compute baselines, including EDSR (16.25\,GFLOPs, 25.43\,dB), SwinIR-light (8.72\,GFLOPs, 25.30\,dB), SRFormer-light (8.25\,GFLOPs, 25.23\,dB), and SR3 (4{,}376\,GFLOPs, 25.57\,dB) in PSNR.
SwinIR-Medium attains the highest raw distortion metrics at 26.52\,dB PSNR and 0.7105 SSIM, but does so with 107.11\,GFLOPs, which is about $8.0\times$ the compute of \ours{} Exit~3 (13.34\,GFLOPs). Thus, even in this harder $\times$8 setting, \ours{} remains competitive with a much larger and much more expensive transformer while delivering better quality than all baselines in the low- to moderate-compute regime.
Adaptive routing remains effective, with 32\% E1, 64\% E2, and only 4\% E3 exits, achieving 26.14\,dB at reduced compute and the second-best LPIPS (0.3532), close to SR3's 0.3504.
This suggests that as the ambiguity of SR increases at $\times$8, larger-capacity models help, but the flow-matching formulation still scales well: each LR pixel corresponds to 64 HR pixels, yet \ours{} preserves strong reconstruction quality with a single forward pass rather than multi-step denoising.

Figure~\ref{fig:qualitative_x8} confirms the quantitative findings visually.
At $\times$8, the lighter regression baselines (EDSR, SwinIR-light, SRFormer-light) produce heavily smoothed outputs that lose fine cellular detail.
SwinIR-Medium and \ours{} recover substantially sharper structures, while SR3 restores some texture through iterative denoising but at much higher compute.
\ours{} remains visually competitive across all three organs, with particularly clear kidney boundaries, while SwinIR-Medium is the strongest fixed regression baseline in this harder setting.

\section{Conclusion}
\label{sec:conclusion}

We presented \ours{}, an adaptive-depth flow matching framework for efficient super-resolution that combines three novel components: (1)~single-step flow matching in rearranged space, where $t{=}0$ mixing is essential ($-$1.54\,dB without it); (2)~a graduated hybrid backbone with quality-aware routing (1.90M parameters, ${\sim}$6K classifier) achieving 33\% compute savings at 0.12\,dB cost; and (3)~comprehensive validation including downstream nuclei segmentation, per-organ analysis, cross-organ generalization, and $\times$8 scaling.

On multi-organ histopathology $\times$4 SR, \ours{} achieves 31.84\,dB PSNR at full depth (13.3\,GFLOPs) and 31.72\,dB with adaptive routing (8.9\,GFLOPs, 33\% savings). A much larger SwinIR-Medium reaches similar PSNR only at 107.1\,GFLOPs, while \ours{} retains the best SSIM among the x4 methods.
The method generalizes to held-out colon tissue with minimal quality degradation ($-$0.02\,dB), and at $\times$8 remains competitive with a stronger SwinIR-Medium baseline while outperforming all lighter regression baselines.
Downstream nuclei segmentation reveals that detection F1 has saturated across learned SR methods ($p > 0.05$), making computational efficiency an important factor for clinical deployment.
Although generative SR methods can in principle hallucinate fine-grained structures, \ours{} may reduce this risk through its distortion-oriented objective (L1 + SSIM, no adversarial or perceptual loss), single-step inference that avoids iterative error accumulation, and downstream validation showing that nuclei detection F1 is statistically indistinguishable from HR ground truth.
Wall-clock benchmarks confirm that the GFLOPs savings translate to real speedups: \ours{} Exit~3 processes a tile in 12.4\,ms (80~img/s) at 49\,MB VRAM, compared to 1{,}082\,ms for SR3, 60.1\,ms for SwinIR-light, and 84.1\,ms for SwinIR-Medium; adaptive routing further reduces latency to 8.2\,ms (122~img/s).

\paragraph{Limitations}
While we demonstrate cross-organ generalization (Section~\ref{sec:held_out}) and $\times$8 scaling (Section~\ref{sec:x8}), evaluation remains limited to histopathology; generalization to natural images or other medical modalities (radiology, ophthalmology) is untested.
The exit classifier is trained on oracle labels derived from L1 loss, which may not perfectly correlate with perceptual quality.
Additionally, the current design assumes a fixed number of exits; dynamically adjusting exit placement based on dataset characteristics could further improve efficiency.
The batch-size-1 inference regime, common in clinical deployment, does not fully exploit the adaptive routing's aggregate compute savings; batched inference with mixed-exit assignment would better realize these gains.

\paragraph{Future work}
Promising directions include scaling to larger backbones with more exit points, extending to other imaging modalities (radiology, ophthalmology), and combining exit routing with knowledge distillation~\citep{hinton2015distilling} to further compress early exits.
Exploring patch-level adaptive depth within a single image, where different spatial regions of a slide tile receive different compute budgets, could enable even finer-grained efficiency.
Our downstream nuclei segmentation results suggest that SR-aware loss functions targeting morphological preservation (e.g., detection-based losses) could further improve clinical utility.
Finally, integrating the exit classifier's tissue complexity signal into downstream analysis tasks (e.g., triaging or quality control) could provide additional clinical value beyond super-resolution.

\section*{Acknowledgments}

This work was supported by the Israel Ministry of Science and Technology Digital Pathology grant.

\bibliographystyle{plainnat}
\bibliography{references}

\end{document}